\documentclass[a4paper,fleqn]{cas-sc}

\usepackage[authoryear]{natbib}
\usepackage[caption=false,font=footnotesize,labelfont=rm,textfont=rm,subrefformat=parens]{subfig}
\usepackage{amsmath,amsfonts}
\usepackage{algorithm}
\usepackage[noend]{algpseudocode}
\usepackage{multirow}
\usepackage{array}
\usepackage{booktabs}
\usepackage{threeparttable}
\usepackage{textcomp}
\usepackage{stfloats}
\usepackage{url}
\usepackage{verbatim}
\usepackage{graphicx}
\usepackage{float}

\begin{document}
\shorttitle{}
\shortauthors{Bo Leng et~al.}

\title [mode = title]{Risk-Aware Reinforcement Learning for Autonomous Driving: Improving Safety When Driving through Intersection}

\author[1]{Bo Leng}
\credit{Funding acquisition, Project administration, Resources, Supervision, Writing – review \& editing}
\affiliation[1]{organization={School of Automotive Studies, Tongji University},
                city={Shanghai},
                postcode={201804}, 
                country={China}}
\ead{lengbo@tongji.edu.cn}

\author[1]{Ran Yu}
\credit{Conceptualization, Data curation, Formal analysis, 
Investigation, Methodology, Software, Validation, Visualization, 
Writing – original draft, Writing – review \& editing}
\ead{2433113@tongji.edu.cn}
\author[1]{Zhuoren Li}[orcid=0000-0001-6246-3404]
\cormark[1]
\ead{1911055@tongji.edu.cn}
\credit{Conceptualization, Validation, Software, Supervision, Writing – original draft, Writing – review \& editing}

\author[1]{Wei Han}
\ead{tjhanwei@tongji.edu.cn}
\credit{Writing – review \& editing}
\author[1]{Lu Xiong}
\ead{xiong\_lu@tongji.edu.cn}
\credit{Funding acquisition, Project administration, Supervision, Writing – review \& editing}
\author[2]{Hailong Huang}
\credit{Writing – review \& editing}
\ead{hailong.huang@polyu.edu.hk}

\affiliation[2]{organization={Department of Aeronautical and Aviation
 Engineering, The Hong Kong Polytechnic University},
                city={Hong Kong},
                country={China}}

\cortext[cor1]{Corresponding author. School of Automotive Studies, Tongji University, Shanghai, 201804, China}

\begin{abstract}
Applying reinforcement learning to autonomous driving has garnered widespread attention. However, classical reinforcement learning methods optimize policies by maximizing expected rewards but lack sufficient safety considerations, often putting agents in hazardous situations. This paper proposes a risk-aware reinforcement learning approach for autonomous driving to improve the safety performance when crossing the intersection. Safe critics are constructed to evaluate driving risk and work in conjunction with the reward critic to update the actor. Based on this, a Lagrangian relaxation method and cyclic gradient iteration are combined to project actions into a feasible safe region. Furthermore, a Multi-hop and Multi-layer perception (MLP) mixed Attention Mechanism (MMAM) is incorporated into the actor-critic network, enabling the policy to adapt to dynamic traffic and overcome permutation sensitivity challenges. This allows the policy to focus more effectively on surrounding potential risks while enhancing the identification of passing opportunities. Simulation tests are conducted on different tasks at unsignalized intersections. The results show that the proposed approach effectively reduces collision rates and improves crossing efficiency in comparison to baseline algorithms. Additionally, our ablation experiments demonstrate the benefits of incorporating risk-awareness and MMAM into RL.
\end{abstract}




\maketitle
\section{Introduction}
As one of the most challenging autonomous driving (AD) tasks, navigating through intersection brings inevitable interactions that require a comprehensive consideration of safety, efficiency, timing, and other factors. Traditional rule-based approaches prone to overly conservative or inconsistent driving strategies in this complex condition, making it difficult to pass through safely and efficiently \citep{refn1}.

Recent advancements in reinforcement learning (RL) have highlighted its potential to surpass human driving capabilities, owing to its superior handling of high-dimensional state spaces and adaptability to complex scenarios \citep{ref5}. RL technologies have been extensively explored in various scenarios, including highway, merging, intersection, etc. \citep{ref6, ref7}. RL optimizes policies by maximizing expected rewards. However, classic RL agents may exhibit unsafe behaviors due to a lack of safety consideration. For safety-critical driving tasks, it is essential not only to maximize rewards but also to incorporate safety guarantees to prevent accidents \citep{ref21}. Consequently, the effective application of RL while ensuring safety has become an urgent challenge for promoting AD. 

To address this issue, Safe RL is introduced to ensuring compliance with safety constraints while maximizing rewards. Some approaches enhance safety by introducing safety factors or risk measures into the objective or reward function \citep{ref8,ref9}. The essence of this category is to modify the policy gradient to update the policy in the direction of the feasible region, thus improving the safety of agent \citep{ref10}. While they can enhance safety to some extent, their safety performance deteriorates significantly when confronted with intricate scenarios, resulting in a higher incidence of constraint violations. Other approaches identify unsafe actions during the agent's exploration phase and project them onto a safe set, thereby ensuring fewer or even zero constraint violations during training \citep{ref17, ref18}. While these methods offer better state-wise safety, they typically require accurate system dynamics or other prior knowledge \citep{ref21,ref22}, or are designed for specific applications with constraints of a particular form \citep{ref25, fearfield}. 

In unsignalized intersection scenarios, autonomous vehicles (AVs) encounter traffic from multiple directions, creating potential conflict risks. To represent the surrounding environment, existing studies typically concatenate the AV's state with environmental information into a feature vector and implement policy mapping through a multi-layer perception (MLP) \citep{mlp1, mlp2}. However, these methods face two major challenges: $\textit{dimension sensitivity}$, where traditional MLPs rely on fixed-dimensional feature vector inputs, making it difficult to adapt to the dynamic number of traffic participants, and $\textit{permutation sensitivity}$, where irregular traffic flows lead to abrupt changes in interacting objects and spatial relationships between adjacent time steps, further complicating state characterization and decision-making \citep{fix_dimention}. Some approaches use grid maps \citep{grid} or bird's eye view (BEV) \citep{bev} to represent environmental features and address these issues. Nevertheless, the discretization of grid divisions and downsampling in image encoding can result in the loss of fine-grained information. In \citep{fix_dimention}, an encoding sum and concatenation (ESC) method is proposed, where an MLP maps each surrounding vehicle (SV) to an feature vector, and they are added element-wise to form the surrounding state representation. However, equal-weight summation struggles to filter out information that is strongly correlated with the ego vehicle (EV). Attention mechanism  has been widely used in RL policy construction to capture relationships between features, thereby improving the policy's ability to understand environmental information \citep{ref23, ref30, ref31}. Inspired by the transformer architecture \citep{ref44}, we incorporate the attention mechanism into the network to effectively handle dynamic traffic flow.

The inability to handle the dynamic changes in the number and permutation of surroundings traffic participants may make it difficult for AVs to identify potential risks and adopt unsafe strategies. Moreover, the complex information at these intersections requires the AV to identify pivotal data to ascertain the timing of passage. To improve safety and efficiency for driving through intersection, a risk-aware RL approach is proposed in this paper and the main contributions are summarized as follows:

\begin{itemize}
\item Safe critics are constructed to evaluate driving risk and work in conjunction with the reward critic to update the actor. A Lagrangian relaxation method is incorporated to generate approximate safe actions, which are projected into a feasible safe region with safety iterative correction by cyclic gradient descent.
\item A \textbf{M}ulti-hop and \textbf{M}ulti-layer perception mixed \textbf{A}ttention \textbf{M}echanism (\textbf{MMAM}) integrated into the actor-critic network enables the policy to adapt to dynamic traffic and overcome permutation sensitivity challenges, enhancing scene understanding and improving decision-making timing when navigating intersections.

\item The proposed approach is evaluated through comparative experiments, as well as ablation studies, demonstrating its effectiveness in terms of safety and efficiency.
\end{itemize}

\section{Related Works}
\subsection{Safe RL Methods}
Algorithms based on Safe RL aim to constrain risks within a given threshold or to avert constraint violations. Many approaches use the Lagrange multipliers to transform the constrained optimization problem into an unconstrained one \citep{ref12, ref13}, or utilize the trust region approach, which ensures policy feasibility and stability by constructing an approximate objective at each iteration and restricting the update range \citep{ref14, ref16}. These methods addresses constraints implicitly, but it can only ensure limited safety and is still prone to severe constraint violations in complex scenarios. Another notable branch ensures the safety during training by preventing the agent from exploring risky behaviors. Some algorithms leverage control theory by constraining the agent to a designated feasible region. \citep{ref17} and \citep{ref18} integrate RL with Lyapunov functions and Control Barrier Functions (CBF), respectively, to ensure that state trajectories remain within a safe feasible region. However, these algorithms often require manual specification of safety constraint functions, and accurately determining these functions can be challenging. \citep{ref25} introduce a safety layer that directly modifies the output actions, linearly mapping the original policy to a safe set to ensure safety. However, methods based on linearization assumptions may not accurately represent system dynamics and can lead to approximation errors. 

\subsection{Safe RL Methods for Autonomous Driving}
In safety-critical tasks such as autonomous driving, ensuring safety is essential to prevent catastrophic accidents. Some algorithms enhance safety by incorporating additional safety constraint objectives. For instance, \citep{ref33} evaluate driving risks using probabilistic models that account for position uncertainty and distance-based safety metrics. Similar initiatives have introduced risk assessment and trade-offs within DRL \citep{ref34, ref35}. While the above methods can improve security, it is challenging to avoid the decline in safety performance caused by constraint violations. \citep{ref22} propose a framework based on vehicle trajectory prediction, which incorporates a safety layer to mask unsafe actions. Similarly, \citep{ref23} develop a lightweight safety layer designed to identify and eliminate unsafe actions in advance. \citep{ref36} modify the actions during exploration to obtain approximate safe actions and used them to train safe strategies. Moreover, some algorithms employ reachability analysis to assess the safety of vehicle trajectories. Notably, \citep{ref24} introduce an online reachability analysis algorithm that calculates the occupancy of both the vehicle and surrounding trajectories, ensuring the safety of the vehicle's path.

\section{Preliminaries}
\begin{figure}[!t]
\centering
\includegraphics[width=6.0cm]{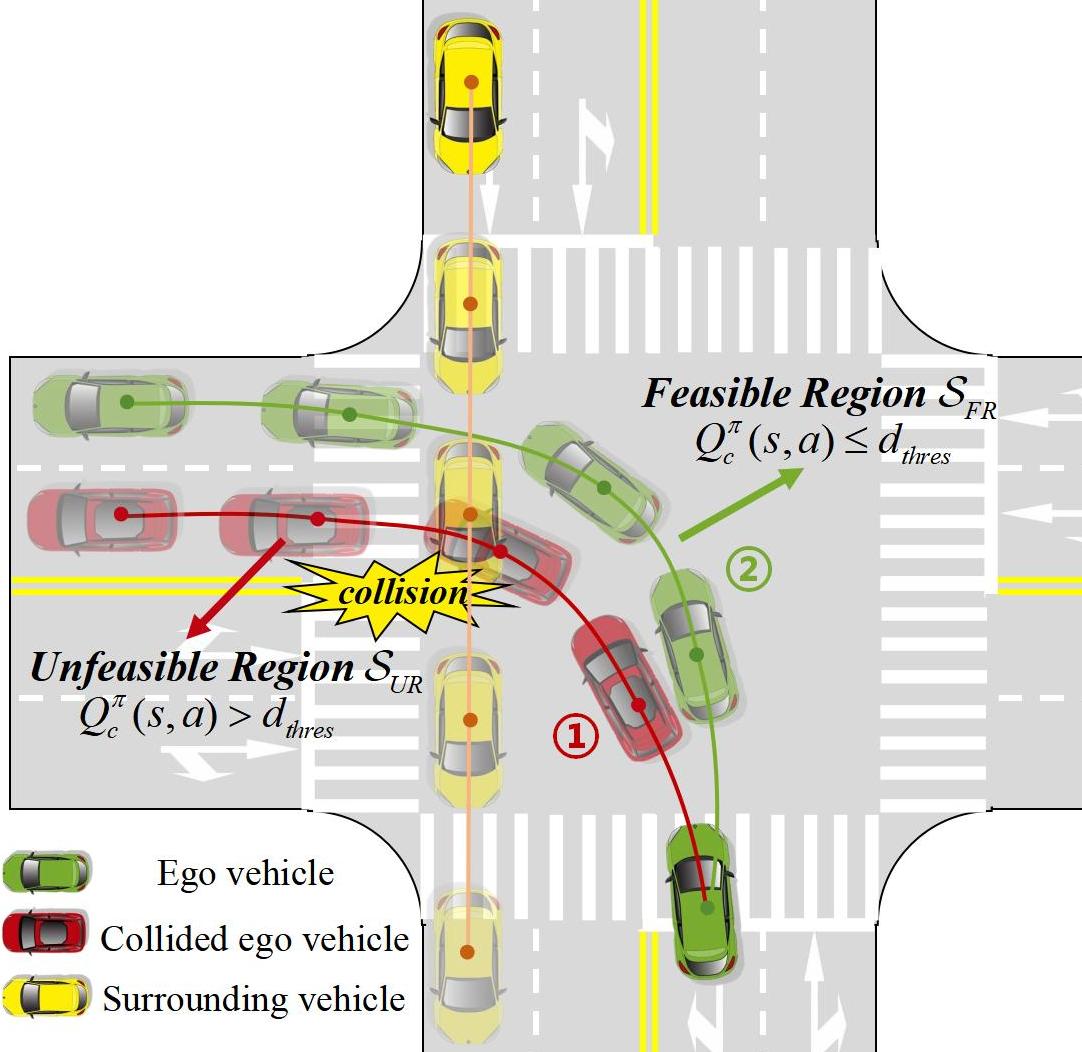}
\caption{Diagram of feasible region $\mathcal{S}_{FR}$ and unfeasible region $\mathcal{S}_{UR}$.}
\label{f1}
\end{figure}
\subsection{Constrained Markov Decision Process}
In this paper, Safe RL is modeled as a Constrained Markov 
Decision Process (CMDP), which extends the standard MDP to a heptuple $(\mathcal S,\mathcal A,\mathcal P, \mathcal R,\mathcal C,\rho,\gamma)$. $\mathcal S$ and $\mathcal A$ are denoted as state space and action space respectively. $P:\mathcal{S}\times\mathcal{A}\times\mathcal{S}\to[0,1]$ is the transition probability function, which represents the system dynamic. $\mathcal{R}:\mathcal{S}\times\mathcal{A}\to\mathbb{R}$ is the reward function. $\mathcal{C}:\mathcal{S}\times{\mathcal{A}}\mapsto[0,+\infty]$ maps the state action transition tuple into a cost value and reflects the constraint violation. $\rho:\mathcal{S}\to[0,1]$ is the initial state distribution and $\gamma$ is the discount factor for future reward and cost. Policy $\pi:\mathcal{S}\to{\mathcal{P}}(\mathcal{A})$ is a map from given states to a probability distribution over action space. In standard MDP, the goal is to optimize the policy by maximizing the agent's cumulative discounted reward:

\begin{equation}
\begin{aligned}[b]
\mathcal{J}_{R}(\pi)=\mathbb{E}_{\tau\sim\pi}\left[\sum_{t=0}^\infty\gamma^t\mathcal{R}(s_t,a_t)\right],
\label{E1}
\end{aligned}
\end{equation} where, $\tau=[s_0,a_0,s_1,\cdots]$, and $\tau\sim\pi$ stands for the stochastic trajectory distribution depended on $s_0 \sim \rho, a_t \sim \pi(\cdot|s_t), s_{t+1} \sim P(\cdot|s_t,a_t)$. CMDP is required to optimize the agent's rewards while guaranteeing that the agent satisfies safety constraints. Hence, CMDP can be formulated as the following constrained optimization problem: 
\begin{equation}
\begin{aligned}[b]
\max_{\pi\in\Pi_S}\mathcal{J}_R(\pi),   
s.t.\mathcal{J}_{C}(\pi)\leq b, 
\label{E2}
\end{aligned}
\end{equation} where, $\Pi_S$ is the set of policies $\pi$, $b \in \mathbb R$ is the constraint threshold. The goal of CMDP is to find a feasible policy set satisfying the cost constraints, i.e., $\Pi_C=\{\pi\in\Pi_S\mid \mathcal{J}_{C}(\pi)\leq b\}$. Similar to the definition of $Q^{\pi}(s,a)$ in standard RL, safe critic $Q^{\pi}_c(s,a)=\mathbb{E}_{\tau\sim\pi}\left[\sum_{t^{\prime}=t}^\infty\gamma^tc_{t^{\prime}}\right] \leq b$ represents the cost-return over a certain time horizon.

\subsection{Safety Correction}

 The objective is to find the feasible policy that satisfies the safety constraint $d_{thres}$ by defining a safe critic $Q^{\pi}(s,a)$ as state-wise constraints:
\begin{equation}
\begin{aligned}[b]
\pi^*=\arg\max_\pi\left[Q^\pi(s,a)\right]\quad\mathrm{s.t.} Q_c^\pi(s,a)\leq d_{thres}.
\label{E3}
\end{aligned}
\end{equation}

As shown in Fig.\ref{f1}, the feasible region $\mathcal{S}_{FR}$: $Q_c^\pi(s,a)\leq d_{thres}$ is defined, and any state $s_t$ within the feasible region $\mathcal{S}_{FR}$ satisfies:
\begin{equation}
\begin{aligned}[b]
\forall s_t\in\mathcal{S}_{FR},s_{t+i}\in\mathcal{S}_{FR},\forall i\in\mathbb{N}_+ .
\label{E4}
\end{aligned}
\end{equation}

Similarly, unfeasible region $\mathcal{S}_{UR}$ is defined as the region where $Q_c^\pi(s,a) > d_{thres}$, and the whole state space $\mathcal{S}_{All} = \mathcal{S}_{UR} \bigcup \mathcal{S}_{FR}$. When the ego vehicle is in $\mathcal{S}_{UR}$ and continues to execute its current policy, there is a significant probability of a collision occurring. To circumvent the aforementioned issue, a natural idea would be to correct the original unsafe actions $a_{old}$ towards the feasible region while attempting to minimize the discrepancy between the new and old actions to the greatest extent possible: 
\begin{equation}
\begin{aligned}[b]
\arg\min\|a_{new}-a_{old}\|\quad\mathrm{s.t.~}Q_c^\pi(s,a)\leq d_{thres}.
\label{E5}
\end{aligned}
\end{equation} 

\begin{figure*}[!t]
\centering
\includegraphics[width=14cm]{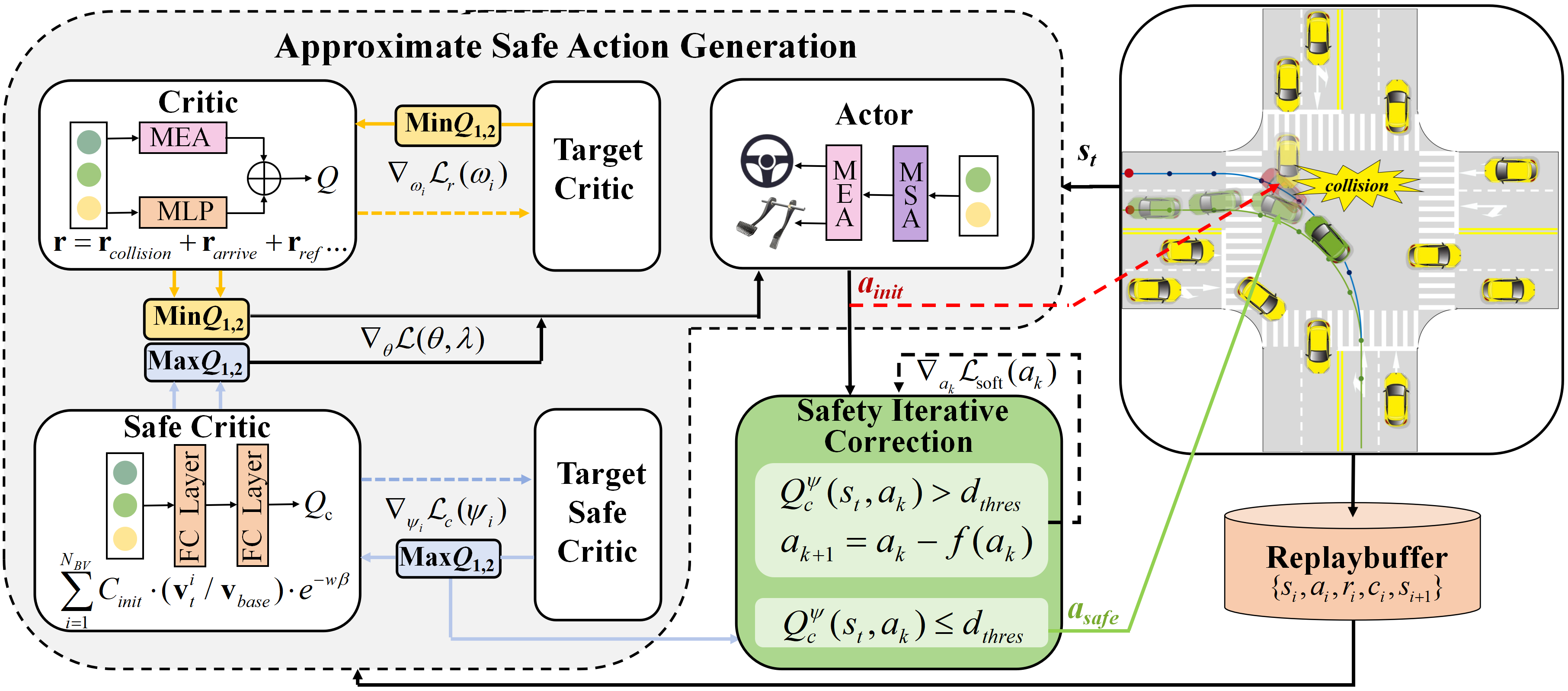}
\caption{Schematic of proposed framework. MEA and MSA stand for multi-head ego-attention and multi-head self-attention, respectively. $f(a_k)=\frac{\eta}{\mathcal N}_k\nabla_{a_k}\mathcal {L}_{\mathrm{soft}}(a_k)$.}
\label{f6}
\end{figure*}
\subsection{Dimension Sensitivity and Permutation Sensitivity}
\subsubsection{Dimension Sensitivity} 
For the observation set \(\mathcal{S}\), it typically contains EV-related information \(\mathcal{S}_{EV} \in \mathbb{R}^{1 \times d_{EV}}\), and SV-related information \(\mathcal{S}_{SV} \in \mathbb{R}^{N_{SV} \times d_{SV}}\),where $d_{EV},d_{SV}$ denote the feature dimension of EV and SVs respectively, and \(N_{SV} \in [1, N] \cap \mathbb{N}\) is the number of observed SVs, which changes dynamically with the traffic flow. That is, \(\mathcal{S} = [\mathcal{S}_{EV}, \mathcal{S}_{SV}]\). Since architectures such as MLP typically require fixed input dimensions, many works use a fixed dimension feature vector, i.e., by specifying a potentially observed number of SVs, \(M_{SV}\). However, if \(M_{SV} < N_{SV}\), the additional vehicles will not be observable, leading to information loss; conversely, when \(M_{SV} > N_{SV}\), it results in information redundancy.

\subsubsection{Permutation Sensitivity}
For the driving policy \(\pi\), we expect it to be \textit{permutation-invariant}. That is, for any two possible permutations \(\zeta_1\) and \(\zeta_2\) of the surrounding traffic participants, the policy \(\pi\) should output the same decision, namely:
\begin{equation}
\begin{aligned}[b]
\begin{aligned}
\pi(\cdot|(\mathcal{S}_{EV}, \mathcal{S}_{SV}^{\zeta_1(1)}, \ldots, \mathcal{S}_{SV}^{\zeta_1(N_{SV})})) = \pi(\cdot|(\mathcal{S}_{EV}, \mathcal{S}_{SV}^{\zeta_2(1)}, \ldots, \mathcal{S}_{SV}^{\zeta_2(N_{SV})})) \quad \forall \zeta_1, \zeta_2 \in \mathfrak{S}_{N_{SV}}.
\end{aligned}
\label{Ep1}
\end{aligned}
\end{equation}

Conversely, if there exist two permutations \(\zeta_1\) and \(\zeta_2\) that make the output of the policy inconsistent, this is called \textit{permutation sensitivity}:
\begin{equation}
\begin{aligned}[b]
\begin{aligned}
\pi(\cdot|(\mathcal{S}_{EV}, \mathcal{S}_{SV}^{\zeta_1(1)}, \ldots, \mathcal{S}_{SV}^{\zeta_1(N_{SV})})) \neq \pi(\cdot|(\mathcal{S}_{EV}, \mathcal{S}_{SV}^{\zeta_2(1)}, \ldots, \mathcal{S}_{SV}^{\zeta_2(N_{SV})})) \quad \exists \zeta_1, \zeta_2 \in \mathfrak{S}_{N_{SV}}.
\end{aligned}
\label{Ep2}
\end{aligned}
\end{equation}

\section{Methodologies}

Directly correcting an unsafe action to a safe one is quite challenging. Hundreds of iterations may be required to project an unsafe action into $\mathcal{S}_{FR}$. Moreover, if the initial action is far from $\mathcal{S}_{FR}$, multiple iterations may still result in the action remaining in $\mathcal{S}_{UR}$. Therefore, we use a Lagrangian relaxation approach to obtain approximate safe action $a_{init}$. Then, Safety Iterative Correction is applied to $a_{init}$ to obtain a feasible safe solution $a_{new}$ when $Q^{\pi}_c(s,a_{init}) > d_{thres}$.

 The overall framework is depicted in Fig. \ref{f6}, which includes two pairs of reward critics $Q^{\omega}_{1,2}$ and target critics $Q^{\omega^{-}}_{1,2}$, as well as two pairs of safe critics $Q^{\psi}_{c_{1,2}}$ and target safe critics $Q^{\psi^{-}}_{c_{1,2}}$, all of which contribute to actor's $\pi_\theta$ policy updates and action risk evaluation. The pseudo-code is shown in Algorithm \ref{A1}.
 
\subsection{Approximate Safe Action Generation}
The initial solution $a_{init}$ is derived through the construction of a Lagrangian function for constrained policy optimization:
\begin{equation}
\begin{aligned}[b]
\begin{aligned}
&\max_{\lambda\geq0}\min_\theta \mathcal {L}(\theta, \lambda)\\=&\max_{\lambda\geq0}\{\min_\theta\mathbb{E}_{s_t\sim\mathcal{D},a_t\sim\pi_\theta}\big[\alpha\log\pi_\theta(a_t\mid s_t)
 - Q^{\omega}(s_t,a_t) + \lambda\big(ReLU(Q_{c}^{\psi}(s_t,a_t) - d_{thres})\big)\big]\}
,\end{aligned}
\label{E11}
\end{aligned}
\end{equation} where $\lambda$ is the Lagrange multiplier, $\alpha$ is the temperature parameter that dictates the relative significance of the entropy term compared to the reward and $\log\pi_\theta(\cdot)$ is entropy of policy $\pi_\theta$. By employing a dual ascent strategy, the algorithm alternately updates the policy and the Lagrange multipliers, thereby gradually converging to the saddle point of the minimax problem:
\begin{equation}
\begin{aligned}[b]
\lambda\leftarrow\lambda+\alpha_\lambda\nabla_\lambda\mathcal{L}(\theta,\lambda),\quad\theta\leftarrow\pi-\alpha_\theta\nabla_\theta\mathcal{L}(\theta,\lambda),
\label{E7}
\end{aligned}
\end{equation} where $\alpha_\lambda$, $\alpha_\theta$ are the step sizes for the parameters $\lambda, \theta$ respectively. Lagrange multiplier functions operate analogously to penalty coefficients, enabling the policy to gradually converge within the constraints.

A dual-critic is employed in both reward critic and safe critic, to mitigate positive bias during the policy improvement step to address the overestimation issue. The larger $Q_{c}^{\pi}(s,a) = \max_{i=1,2} Q_{c,i}^{\pi}(s,a)$ is selected to reduce the risk of underestimation. Although it may lead to an overestimation of the $Q_{c}^{\pi}(s,a)$, it effectively increases the safety margin for overall reliability. Therefore, reward critic network $Q^{\omega}$ and safe critic network $Q^{\psi}_{c}$ can be updated by as follows:
\begin{equation}
\begin{aligned}[b]
\begin{aligned}&\mathcal{L}_{r}(\omega_{i})=\frac{1}{2}\left\{Q_{i}^{\omega}(s_{t},a_{t})-\Big(r(s_{t},a_{t})+\gamma V_{r}^{\omega^{-}}(s_{t+1})\Big)\right\}^{2}\\&with\quad V_{r}^{\omega^{-}}(s_{t+1}) =\min_{j=1,2}Q_{r,j}^{\omega}(s_{t+1},a_{t+1}) -\alpha\log\pi(a_{t+1}| s_{t+1}),\end{aligned}
\label{E12}
\end{aligned}
\end{equation}
\vspace{-15pt}
\begin{equation}
\begin{aligned}[b]
\begin{aligned}&\mathcal{L}_{c}(\psi_{i})=\frac{1}{2}\left\{Q_{c,i}^{\psi}(s_{t},a_{t})-\Big(c(s_{t},a_{t})+\gamma V_{c}^{\psi^{-}}(s_{t+1})\Big)\right\}^{2}\\&with\quad V_{c}^{\psi^{-}}(s_{t+1})=\max_{j=1,2}Q_{c,j}^{\psi}(s_{t+1},a_{t+1}), \end{aligned}
\label{E13}
\end{aligned}
\end{equation}

The temperature parameter can be adjusted adaptively. As the policy becomes more definitive in the later stages of training, the exploration capability can be appropriately reduced. Specifically, the update of the temperature parameter is guided by the following optimization objective:
\begin{equation}
\begin{aligned}[b]
\mathcal{L}(\alpha)=\mathbb{E}_{(s_t,a_t)\sim\rho_\pi}[-\alpha\log\pi(a_t\mid s_t)-\alpha\mathcal{H}_0],
\label{E14}
\end{aligned}
\end{equation} where $\rho_\pi$ is the state distribution under policy $\pi$ and $\mathcal{H}_0$ is the target entropy.

\subsection{Safety Iterative Correction}
Based on (\ref{E5}), we refine the initial solution to ensure safety using the following soft loss function:
\begin{equation}
\begin{aligned}[b]
\begin{aligned}
\mathcal {L}_{\mathrm{soft}}(a_k)&=\frac{1}{2}\|a_{k}-a_{init}\|^2+\lambda_{a}\big(ReLU(Q_{c}(s,a_{k}) - d_{thres})\big),
\end{aligned}
\label{E8}
\end{aligned}
\end{equation} where $k$ denotes the $k$-th iteration and $\lambda_a$ represents the coefficient of the constraint. The corresponding gradient for $k$-th action $a_k$ is:
\begin{equation}
\begin{aligned}[b]
\begin{aligned}
\nabla_{a_k}\mathcal {L}_{\mathrm{soft}}(a_k)&=a_{k}-a_{init}+\lambda_{a}\frac{\partial\big(ReLU(Q_{c}(s,a_{k}) - d_{thres})\big)}{\partial a_k}. 
\end{aligned}
\label{E9}
\end{aligned}
\end{equation}

Ultimately, the gradient descent method is utilized to update the action $a_k$: 
\begin{equation}
\begin{aligned}[b]
a_{k+1}=a_k-\frac{\eta}{\mathcal N}_k\nabla_{a_k}\mathcal {L}_{\mathrm{soft}}(a_k),
\label{E10}
\end{aligned}
\end{equation} where hyper-parameter $\eta$ determines the update magnitude for each iteration and $\mathcal{N}_k=\|\nabla_{a_k}\mathcal {L}_{\mathrm{soft}}(a_k)\|_\infty$ is the scaling factor normalize the gradients on $a_k$. $a_k$ is expected to converge to the optimal action $a_k^*$ as $k\rightarrow\infty$. Due to time constraints, a more practical approach is to set a maximum iteration limit $N_{iter}$. If this limit is reached without satisfying the constraints, the iteration will be terminated. 

\begin{figure}[!t]
    \centering
    \subfloat[]{%
        \includegraphics[width=3.5cm]{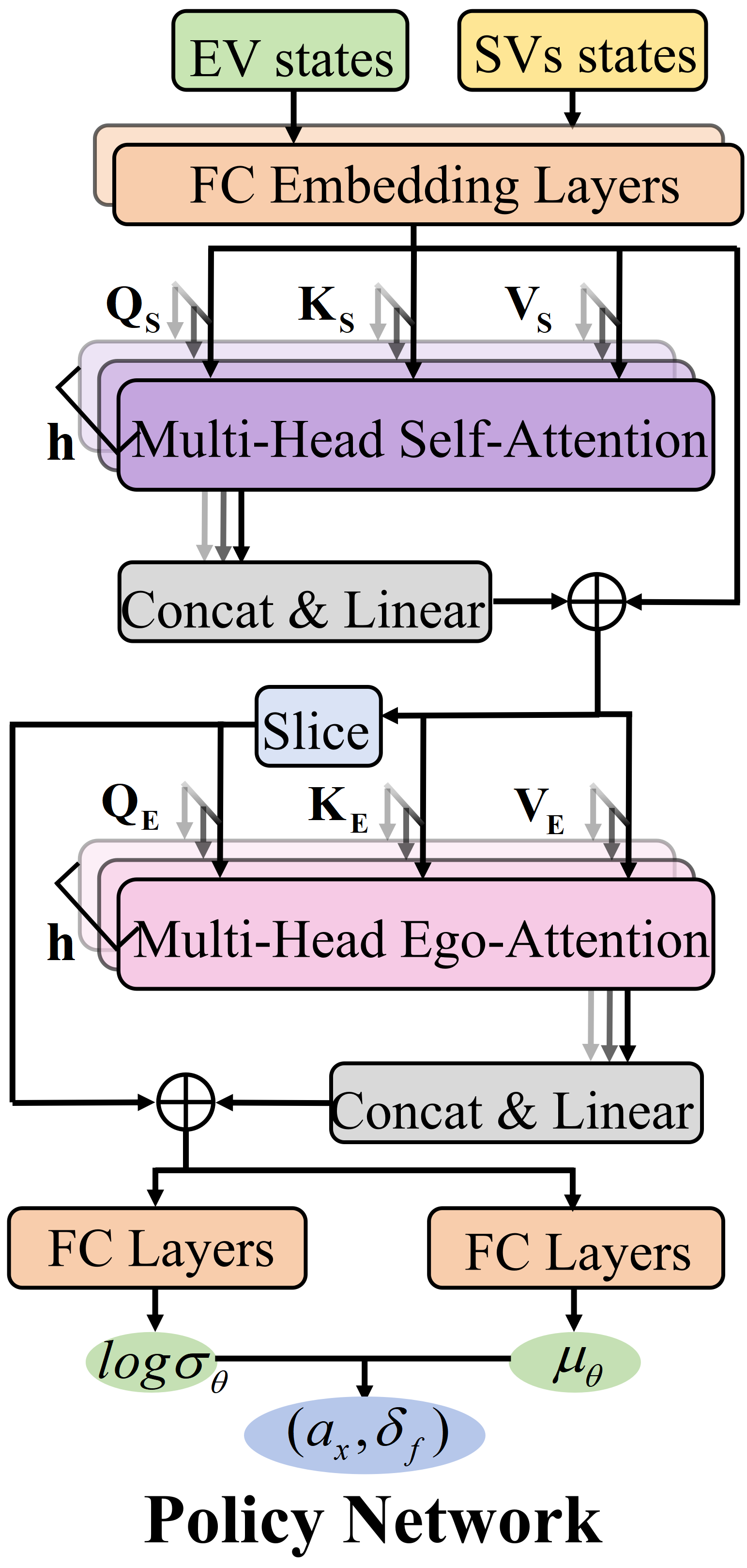}
        \label{a_a}
    }\subfloat[]{%
        \includegraphics[width=5.0cm]{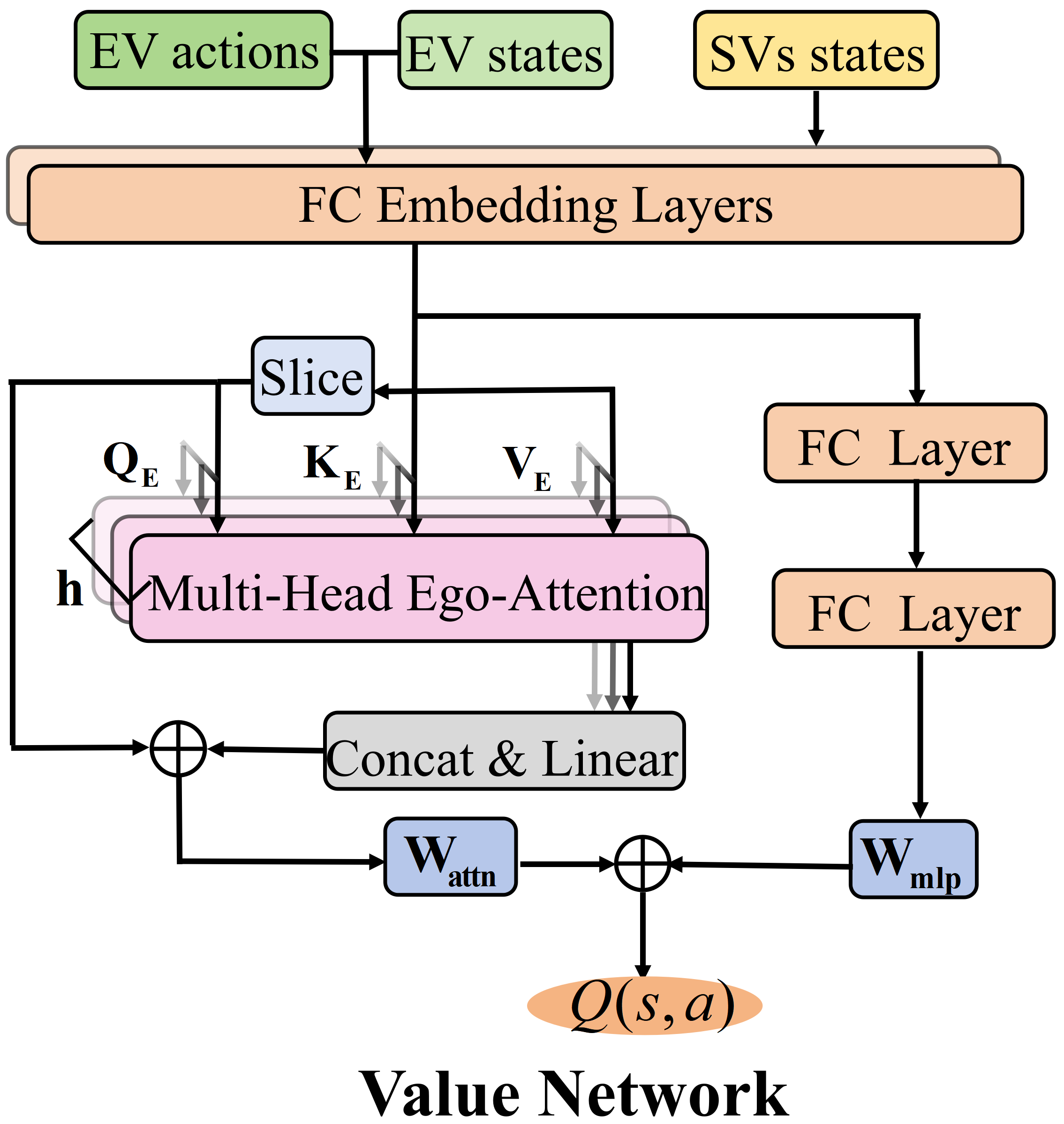}
        \label{a_b}
    }
    \caption{Multi-hop and MLP-mixed Attention Mechanism (MMAM).}
    \label{f5}
\end{figure}
    
\subsection{Attention embedded Actor-Critic Network}
    
To tackle the dimension and permutation sensitivity, MMAM is incorporated into the actor-critic network, as shown in Fig. \ref{f5}, which enhances the extraction of scene information and improves the scene comprehension capabilities of EV, allowing them to focus on potential risks more effectively. 

With regard to the detailed architecture of the policy network, $\mathcal S_{EV} \in \mathbb{R}^{1\times d_{EV}}$ and $\mathcal S_{SV} \in \mathbb{R}^{N_{SV}\times d_{SV}}$ are processed through their respective fully connected embedding layers, then concatenated and mapped to the latent input matrix $\mathbf{Z}_{1} \in \mathbb{R}^{\mathbf{N}\times d}$, where $d$ represents the hidden size of networks, $\mathbf{N}$ is the total number of vehicles. Then, $\mathbf{Z}_{1}$ can be further transformed into query, key, and value matrices $\mathbf{Q}_{\mathbf{S}}$, $\mathbf{K}_{\mathbf{S}}$, $\mathbf{V}_{\mathbf{S}}\in \mathbb{R}^{\mathbf{N}\times d}$, respectively, using a linear transformation operator $\mathcal{T}\in\mathbb{R}^{d\times d}$. Multi-head attention is  subsequently employed to focus on different parts of the $\mathbf{Z}_{1}$ and the outputs of each head are then merged and transformed back to their original dimensions, as shown below:

\begin{equation}
\begin{aligned}[b]
&\mathrm{MultiHead}(\mathbf{Q,K,V)} =\mathrm{Concat}(\mathrm{head}_{1},...,\mathrm{head}_{\mathrm{h}})\mathbf{W}^{O}\\&\mathrm{where~head}_{i}=\mathrm{Attention}(\mathbf{Q}\mathbf{W}_{i}^{Q},\mathbf{K}\mathbf{W}_{i}^{K},\mathbf{V}\mathbf{W}_{i}^{V})\\
&\mathrm{Attention}(\mathbf{Q,K,V}) =\mathrm{softmax}(\frac{\mathbf{QK}^\top}{\sqrt{d}})\mathbf{V},
\label{AttnE1}
\end{aligned}
\end{equation} where parameter matrices   $\mathbf{W}_{i}^{Q},\mathbf{W}_{i}^{K},\mathbf{W}_{i}^{V}\in\mathbb{R}^{d\times d/h}$ and $\mathbf{W}^{O}\in\mathbb{R}^{d\times d}$. Subsequently, the output $\mathbf{Z}'_{1}$ is augmented with a residual connection, sliced, into and then linearly transformed to form the query $\mathbf{Q}_{\mathbf{E}}\in\mathbb{R}^{1\times d}$ for ego-attention, in order to capture the interaction between EV and SVs. Ego-attention is a variant of self-attention wherein the query $\mathbf{Q}$ solely contains EV's features. This configuration establishes a 2-hop attention structure in conjunction with self-attention, facilitating the iterative integration and extraction of additional feature information through the sequential processing of the query and the latent matrix \citep{ref45}. The parameters of the embedding and attention layers are independent of $\mathbf{N}$, allowing the models to adapt to dynamic input. Meanwhile, since the final result is the dot product of values and key similarities, the model is permutation invariant.

The value network incorporates the EV's actions as input. The input feature matrix $\mathbf{Z}_{2}\in \mathbb{R}^{(N+2)\times d}$ is processed through both the ego-attention branch and the MLP branch, thereby enabling the model to capture relationships between the EV and SVs, as well as global information about states and environment.  Let $\mathbf {Y}_{\mathrm{attn}},\mathbf {Y}_{\mathrm{mlp}}\in \mathbb{R}^{1\times d}$ represent the outputs of the attention and MLP branches, respectively. A weighted sum of these outputs is computed using learnable weight vectors $\mathbf {W}_{\mathrm{attn}}, \mathbf {W}_{\mathrm{mlp}}\in \mathbb{R}^{d\times 1}$, yielding the final $Q$-value:
\begin{equation}
\begin{aligned}[b]
Q(s,a)=\mathbf {Y}_{\mathrm{attn}}\cdot \mathbf {W}_{\mathrm{attn}} + \mathbf {Y}_{\mathrm{mlp}}\cdot \mathbf {W}_{\mathrm{mlp}}. 
\label{AttnE2}
\end{aligned}
\end{equation}

\floatname{algorithm}{Algorithm} 
\renewcommand{\algorithmicrequire}{\textbf{Initialize:}} 
\renewcommand{\thealgorithm}{1}
    \begin{center}
    \begin{minipage}{.7\linewidth}  
    \begin{algorithm}[H]
        \caption{Risk-Aware Soft Actor-Critic} 
        \begin{algorithmic}[1]
        \Require parameters $\omega_1, \omega_2$, $\psi_1, \psi_2$, $\theta$,  $\lambda$; $\omega_1^{-}\leftarrow\omega_1, \omega_2^{-}\leftarrow\omega_2$, $\psi_1^{-}\leftarrow\psi_1, \psi_2^{-}\leftarrow\psi_2$; replay buffer $\mathcal{D}\leftarrow\emptyset$; learning rate $\alpha_r, \alpha_c, \alpha_\theta, \alpha_\lambda, \beta_\alpha$.
        \For{each episode $e$}
            \For{each time-step $t$}
            \State Get state $s_t$ and select action: $a_t\sim\pi(a_t|s_t)$.
            \State Get safe critic value:
             $Q_{c}^{\psi}(s_t,a_t) = \max_{i=1,2} Q_{c,i}^{\psi}(s_t,a_t)$.
            \If {$Q_{c}^{\psi}(s_t,a_t) > d_{thres}$}
                \For {each iteration $k=1\to N_{iter}$}
                    \State $a_{k}\leftarrow a_{k}-\frac{\eta}{\mathcal N}_k\nabla_{a_k}\mathcal {L}_{\mathrm{soft}}(a_k)$.
                    \If {$Q_{c}^{\psi}(s_t,a_k) \leq d_{thres}$}
                        \State Break.
                    \EndIf
                \EndFor
            \EndIf
            \State Execute $a_t$, receive next state $s_{t+1}$, reward $r_t$ and \State cost $c_t$. Store the transition $(s_t, a_t, r_t,c_t,s_{t+1})$ in \State replay buffer $\mathcal{D}$.
            \EndFor
            \For{each training epoch $i$}
                \State Sample $N$ transitions from replay buffer $\mathcal{D}$.
                \State Update the critic network and safe critic network:
                \State $\omega_j\leftarrow\omega_j-\alpha_r{\nabla}_{\omega_j}\mathcal {L}_r(\omega_j),\mathrm{for} j\in\{1,2\}$,
                \State $\psi_j\leftarrow\psi_j-\alpha_c{\nabla}_{\psi_j}\mathcal {L}_c(\psi_j),\mathrm{for} j\in\{1,2\}$.
                \State Update actor network and Lagrange multiplier:
                \State $\theta\leftarrow\theta-\alpha_{\pi}{\nabla}_{\theta}\mathcal{L}(\theta, \lambda)$, $\lambda\leftarrow\lambda+\alpha_{\lambda}{\nabla}_{\lambda}\mathcal{L}(\theta, \lambda)$.
                \State Update temperature parameter:
                \State $\alpha\leftarrow\alpha-\beta_{\alpha}{\nabla}_{\alpha}\mathcal{L}(\alpha)$.
                \State Soft update the target network $\omega_j^-$, $\psi_j^-$.
            \EndFor
        \EndFor
        \end{algorithmic}
        \label{A1}
    \end{algorithm}
    \end{minipage}
    \end{center}

\section{Implementation}
\subsection{Environment Settings}
\label{subsec:env_set}
\begin{figure}[!t]
    \centering
    \subfloat[]{%
        \includegraphics[width=0.25\textwidth]{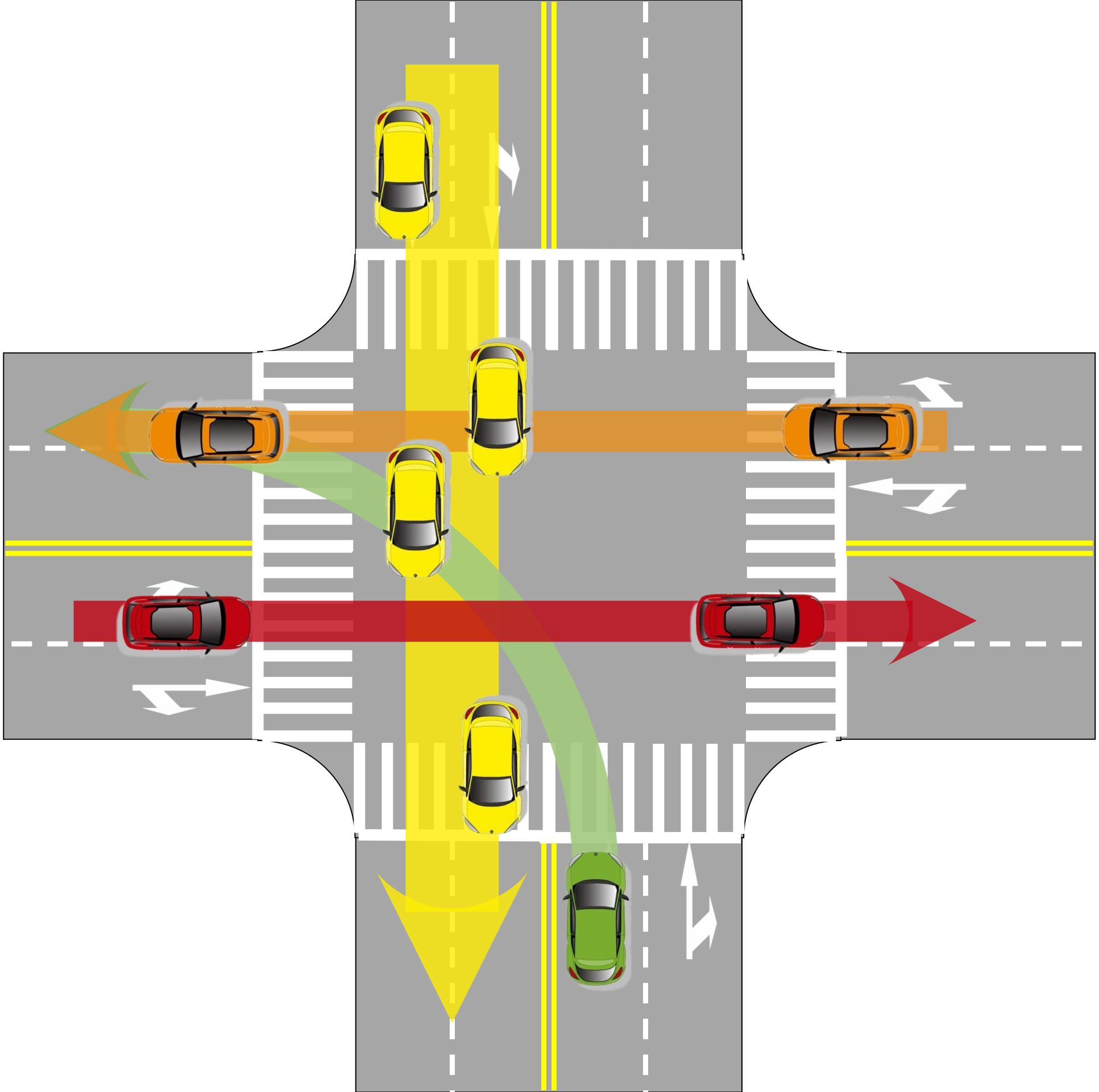}
        \label{task_a}
    }\subfloat[]{%
        \includegraphics[width=0.25\textwidth]{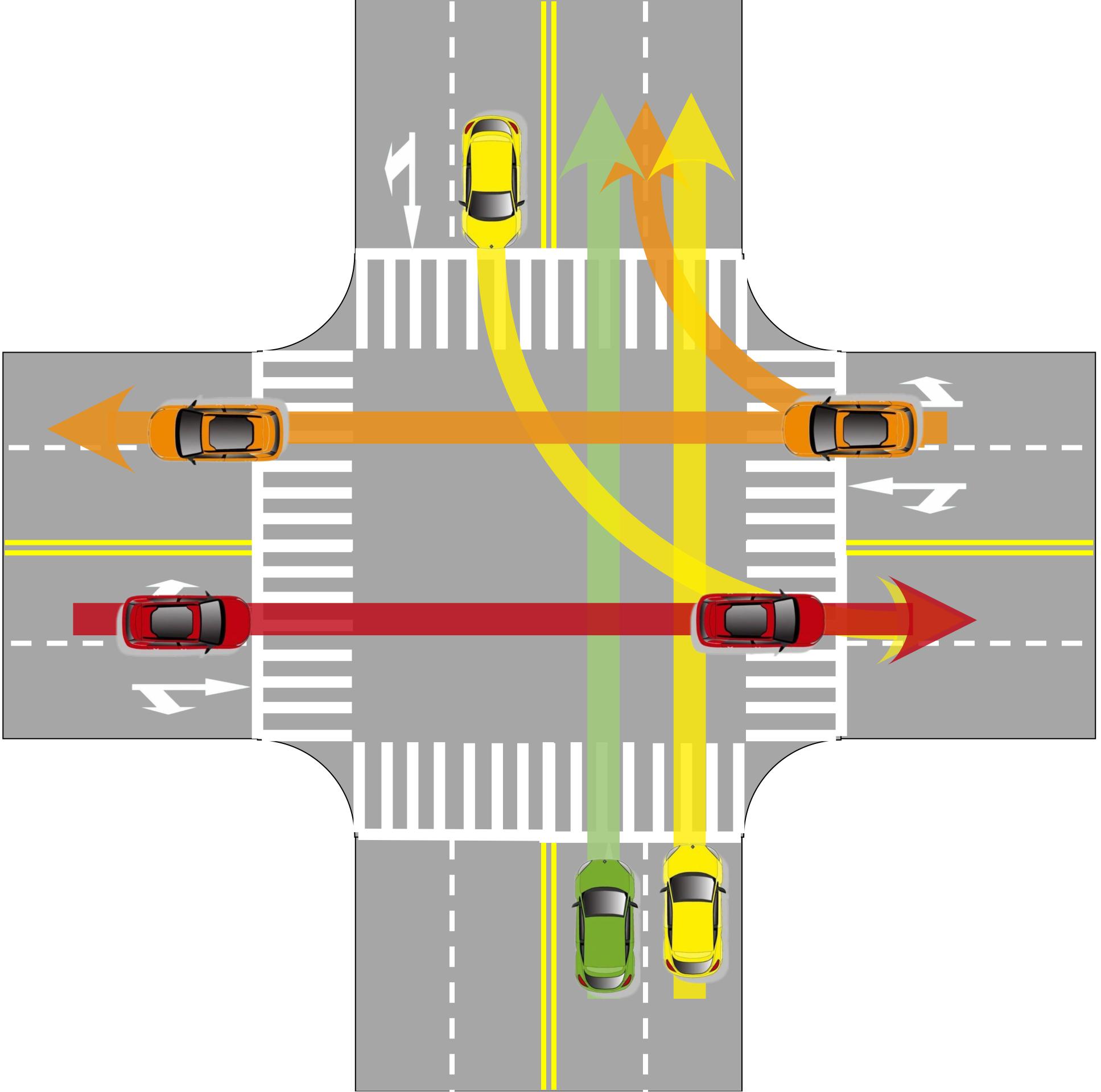}
        \label{task_b}
    }\subfloat[]{%
        \includegraphics[width=0.25\textwidth]{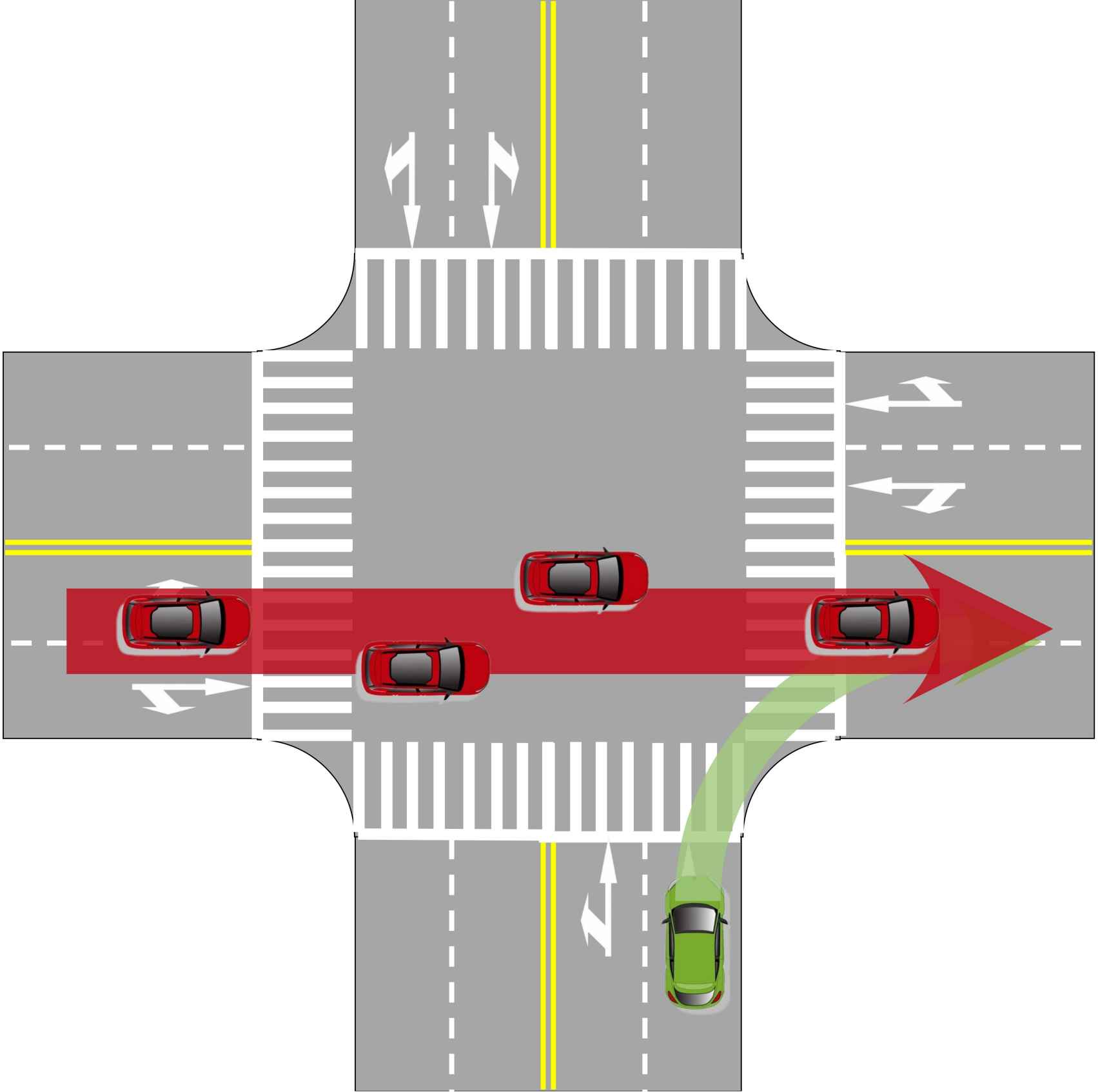}
        \label{task_c}
    }
    \caption{Driving tasks and main conflicts at unsignalized intersection. (a) LT task, EV primarily encounters conflicts with oncoming traffic and some crossing traffic. (b) GS task with mixed traffic flow. (c) RT task with crossing traffic, EV needs to perform a right merge.}
    \label{f11}
\end{figure}

We constructed a bidirectional four-lane intersection scenario based on Highway-Env \citep{ref40} and designed three driving tasks: left-turn (LT), go-straight (GS), and right-turn (RT). As graphed in Fig. \ref{f11}, to avoid sparse traffic flow caused by the random generation of SVs, which would simplify the task to a path-following problem, we specially design the difficulty of the driving task. To reduce collisions from random SV generation, we applied an improved intelligent driver model (IDM) \citep{idm} strategy that follows traffic rules. Each SV predicts its heading and position for the next 2 seconds, yielding to potentially colliding vehicles based on established road priorities. Each time the scenario is reset, 10 SV will be initialized and generated. The initial velocity of each SV is randomly generated within the range of [6 m/s, 10 m/s]. The minimum distance between vehicles is 15 m, and vehicles that do not meet this requirement will be removed. The EV is initialized with a random velocity and positioned in a lane where no collisions will occur at that moment. The simulation frequency $f_{s}$ is 15 Hz, with the policy execution frequency $f_{\pi}$ set to 5 Hz during training and 10 Hz during testing. The maximum length of each episode is 125 time steps (25s).

\subsection{CMDPs Design}
\begin{figure}[!t]
\centering
\includegraphics[width=6.0cm]{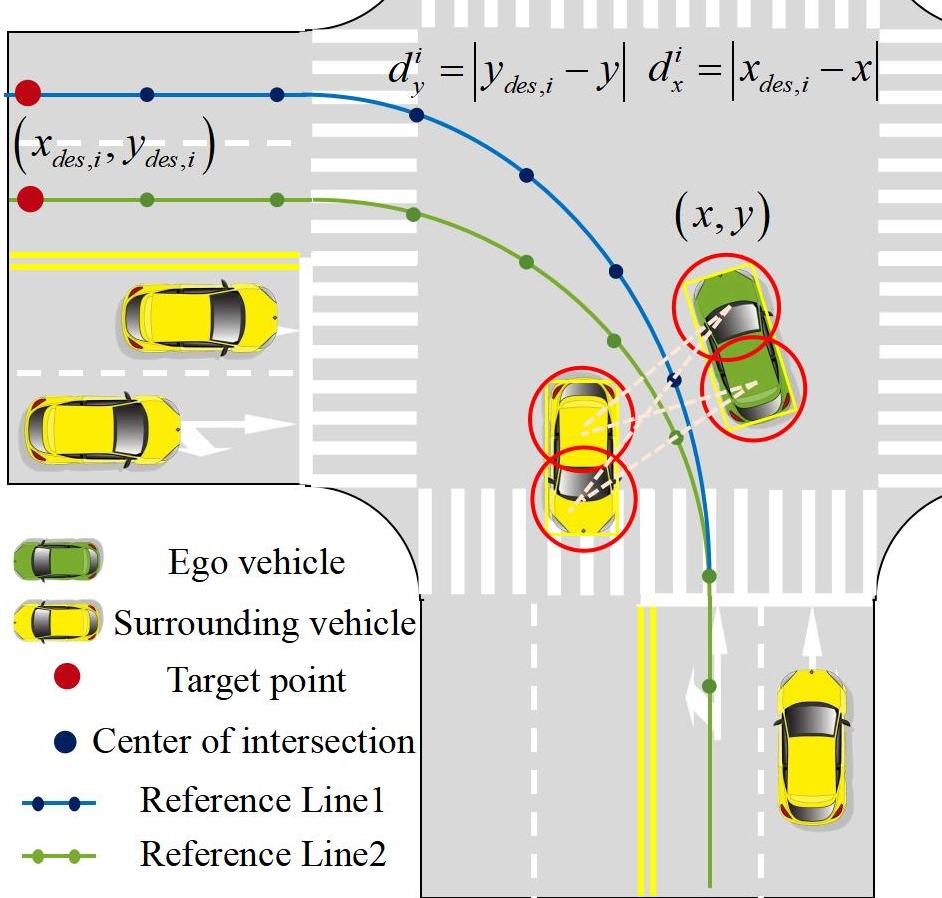}
\caption{Design of the scenario. $d_x^i,d_y^i$ are the relative distances of the EV to the target point along the x-axis and y-axis. $i =1,2$ represents the indices of reference line1 and reference line2.}
\label{f3}
\end{figure}
\subsubsection{observation space and action space}
The observation space is constituted by two components: EV's states $\mathcal{S}_{EV}$ and SVs's states $\mathcal{S}_{SV}$, namely: $\mathcal{S}$ = $[\mathcal{S}_{EV},\mathcal{S}_{SV}^1,...,\mathcal{S}_{SV}^{N_{SV}}],$ where ${S}_{EV}$ = $[\mathbb{I}_{veh}$,$x$,$y$,$v_{x}$,$v_{y}$,$\phi$,$\omega$,$d_{veh}$,$d_{des}]$, ${S}_{SV}^{j}$ = $[\mathbb{I}_{veh}$,$\Delta x$,$\Delta y$,$\Delta v_{x}$,$\Delta v_{y},\phi]$. The $N_{SV}$ vehicles situated in the closest proximity to EV are selected, with the distance measured from 70 $m$ in front of the EV to 30 $m$ behind it. In the Highway-Env simulator, the fixed-dimensional constraint of the observation space inherently contradicts the dynamic nature of traffic scenarios. To address this limitation, we construct an observation tensor with a constant dimension by predefining a maximum number of SVs \( M_{SV} > N_{SV} \in [1, N] \cap \mathbb{N} \), ensuring that observation redundancy is always present. Additionally, an indicator function \( \mathbb{I}_{veh} \in \{0, 1\} \) is used to indicate whether a vehicle is actually observed (\( \mathbb{I}_{veh} = 1 \)) or is a redundant vehicle (\( \mathbb{I}_{veh} = 0 \)). To mitigate the impact of redundant features on the network, we apply masks to both the feature and attention layers: redundant parts of the feature vector are zero-padded, and when computing the attention weights, a large negative bias (-1e9) is applied to the $\mathbf{QK^\top}$ similarity of the redundant vehicles. This takes advantage of the exponential decay property of the softmax function, causing the normalized weight to approach zero. $\Delta x,\Delta y,\Delta v_{x},\Delta v_{y}$ represent the position and velocity of the SVs relative to the EV. $\phi$ and $\omega$ is the heading angel and yaw rate of vehicle, respectively. As illustrated in Fig. \ref{f3}, $d_{des}= \min_{i = 1,2} ({d_x^i} + {d_y^i})$ is the shortest Manhattan distance from EV to the target points. $d_{veh}=d_{min}-r_{EV}-r_{SV}$ is the shortest distance from the EV to SVs, where $d_{min}$ represents the minimum distance among the four inter-center distances computed between the circular, $r_{veh}=\sqrt{(l_{veh}/4)^2+(w_{veh}/2)^2},veh\in[EV,SV]$, $l_{veh}$ and $w_{vhe}$ are the length and width of the vehicle. We directly control the vehicle's front wheel steering angle $\delta_f$ and longitudinal acceleration $a_{x}$, so the continuous action can be expressed as $a = [ a_{x}, \delta_f]$.

\subsubsection{reward function}

Our reward function is comprised primarily of sparse $\mathbf{r}_{sparse}$ and dense rewards $\mathbf{r}_{dense}$. The sparse rewards, which are used to penalize collisions and encourage reaching the target points, are illustrated as follows:
\begin{equation}
\begin{aligned}[b]
\begin{aligned}
&\mathbf{r}_{sparse}=\mathbf{r}_{collision}+\mathbf{r}_{arrive\_goal},\\
&\mathbf{r}_{collision}=-50\cdot\mathbb {I}_{collision},\\
&\mathbf{r}_{arrive\_goal}=100\cdot\mathbb {I}_{arrive\_goal}.
\end{aligned}
\label{Er_1}
\end{aligned}
\end{equation}

To determine the dense reward function, we consider factors such as reference line information, action smoothness, the distance to destination and safety distance:
\begin{equation}
\begin{aligned}[b]
\begin{aligned}
&\mathbf{r}_{dense}=\frac{2}{1+\mathbf{r}_{ref}}+\mathbf{r}_{smooth}+\mathbf{r}_{des}+\mathbb{I}_{RS}\mathbf{r}_{safe},\\
&\mathbf{r}_{ref}=\max_{i=1,2}(\mathbf{x}_{t,i}^\mathrm{ref}-\mathbf{x}_{t})^\top   Q(\mathbf{x}_{t,i}^\mathrm{ref}-\mathbf{x}_{t}),\\
&\mathbf{r}_{act}=-(a_t^\top R_u a_t + R_{\Delta}\Delta a_t),\\
&\mathbf{r}_{des}=- d_{des}^2 = -[\mathop {\min }\limits_{i = 1,2} ({d_x^i} + {d_y^i})]^2,\\
&\mathbf{r}_{safe}=\begin{cases} 0.0 & \text{if } d_{veh} > 0.5, \\ 
-(1.0 - d_{veh}) & \text{if } 0.2 < d_{veh} \leq 0.5, \\ 
-3 \times (1.0 - d_{veh}) & \text{if } d_{veh} \leq 0.2. \end{cases}\\
\end{aligned}
\label{Er_2}
\end{aligned}
\end{equation} where $\mathbf{x}_t^\mathrm{ref} = [x^\mathrm{ref},y^\mathrm{ref},v_{x}^\mathrm{ref},0,\phi^\mathrm{ref},0]^\top$. In order to guarantee safety, vehicles should adhere to a speed limit of 30 or 40 km/h when approaching and traversing intersections. Consequently, $v_{x}^\mathrm{ref}$ is set at 9 m/s. For $\mathbf{r}_{ref}$, calculate both reference lines simultaneously and select the maximum value to encourage the EV to stay close to the reference line. $\mathbf{r}_{act}$ is employed to encourage the EV to save energy and smooth the trajectory. The weight coefficient matrix or vector $Q=diag(400.0, 400.0, 20.0, 20.0, 2.0, 0.5)$, $R_{a}=diag(0.05, 0.02)$, $R_{\Delta}=[0.10,0.10]$. $\mathbf{r}_{safe}$ is used exclusively in the baseline algorithm based on reward shaping, i.e., when $\mathbb{I}_{RS}=1$. 

\subsubsection{cost function}

To evaluate the existing risk of collision and facilitate the autonomous vehicles' capacity to proactively anticipate potential collision threats, we proposes a cost function based on vehicle trajectory prediction. The predicted positions and headings $\{\mathbf X, \mathbf Y, \mathbf \Phi\}_{EV,SV}$ derived from vehicle dynamic model $\mathcal{F}_{EV}$ and kinematic model $\mathcal{F}_{SV}$ (Sec. \ref{subsec:veh}) employed in the calculation of the vehicle's corner points $\mathcal{V}_{EV,SV}$. The separating axis theorem (SAT) is then employed to detect collisions \citep{ref42}. Considering that the uncertainty of trajectory predictions increases with distance, vehicle expansion coefficient $\beta\in[1.20,1.50]$ is introduced to enlarge the vehicle's rectangular bounding box. The process of the cost function is illustrated in Algorithm 2.

\subsection{Vehicle Model}
\label{subsec:veh}
We use a kinematic model to describe the motion of SVs, while employing a more precise dynamic model to simulate the motion of the EV. Due to the singularity of conventional dynamic models at low speed, we introduce a discrete dynamic bicycle model inspired by the backward Eulerian method that is feasible at any low speed (i.e., less than 15 m/s) \citep{ref41}. This model has been demonstrated to be numerically stable and to have lower prediction errors compared to kinematic models. The transition model of EV and SVs are depicted as follows:
\begin{subequations}
\begin{equation}
\begin{aligned}[b]
\mathbf{x}_{t+1} = \mathcal{F}_{EV} (\mathbf{x}_t,\mathbf{u}_t), \mathbf{x}_{t+1}^{j}=\mathcal{F}_{SV}(\mathbf{x}_{t}^{j}),
\label{E5_1a}
\end{aligned}
\end{equation}
\vspace{-20pt}
\begin{equation}
\begin{aligned}[b]
\mathcal{F}_{EV}=\begin{bmatrix}x+T_s(v_{x}\cos\phi-v_{y}\sin\phi)\\y+T_s(v_{x}\sin\phi+v_{y}\cos\phi)\\v_{x}+T_s(a_x+v_{y}\omega)\\\frac{mv_{x}v_{y}+T_s[(L_fC_f-L_rC_r)\omega-C_f\delta_f v_{x}-mv_{x}^2\omega]}{mv_{x}-T_s(C_f+C_r)}\\\phi+T_s\omega\\\frac{-I_z\omega v_{x}-T_s[(L_fC_f-L_rC_r)v_{y}-L_fC_f\delta v_{x}]}{T_s(L_f^2C_f+L_r^2C_r)-I_zv_{x}}\end{bmatrix},
\label{E5_1b}
\end{aligned}
\end{equation}
\vspace{-15pt}
\begin{equation}
\begin{aligned}[b]
\mathcal{F}_{SV}=\begin{bmatrix}x^j+T_s(v_x^j\cos\phi^j-v_y^j\sin\phi^j)\\y^j+T_s(v_x^j\sin\phi^j+v_y^j\cos\phi^j)\\v_x^j\\0\\\phi^j+T_s\omega_\mathrm{const}^j\end{bmatrix},
\label{E5_1c}
\end{aligned}
\end{equation} 
\end{subequations} where the EV's feature vector $\mathbf{x}_t = [x,y,v_{x},v_{y},\phi,\omega]^\top$, control vector $\mathbf{u}_t=[a_{x},\delta_f]^\top$, SV's feature vector $\mathbf{x}_t^{j}$ is $[x,y,v_{x},v_{y},\phi]^\top$, $x,y$ are the position coordinates of the vehicle's center of gravity, $v_x, v_y$ are the longitudinal and lateral velocities, $\phi$ is the heading angle, $\omega$ is the yaw rate. $L_f, L_r$ are the front and rear wheelbases, respectively. $C_f, C_r$ are the front and rear axle equivalent sideslip stiffness, respectively. $m$ is the mass of vehicle, $I_z$ is the inertia of vehicle's center of gravity. $j\in[1,N_{SV}]$, where $N_{SV}$ represents the total number of SVs. Sampling time $T_s = \frac{1}{f_s}$. The future states of SVs are predicted under constant velocity and yaw rate. Considering actuator saturation, we restrict continuous actions to $a_x\in[-5.0, 5.0]$m/$\text{s}^2$ and $\delta_f\in[0.6,0.6]$rad, respectively.

\floatname{algorithm}{Algorithm} 
\renewcommand{\algorithmicrequire}{\textbf{Initialize:}} 
\renewcommand{\thealgorithm}{2} 
    \begin{center}
    \begin{minipage}{.7\linewidth}  
    \begin{algorithm}[H]
        \caption{Cost function with trajectory prediction}
        \begin{algorithmic}[1]
        \Require EV's velocity $\mathbf v_t$, vehicle expansion coefficient $\beta$, prediction horizon $T$, initial cost value $C_{init}$, base velocity $\mathbf v_{base}$,
        weight coefficient $\mathbf{w}$.
        \State Get the predicted positions and headings of EV and SVs:
        \State $\{\mathbf X, \mathbf Y, \mathbf \Phi\}_{EV}\leftarrow\mathcal{F}_{EV},\{\mathbf X, \mathbf Y,\mathbf \Phi\}_{SV}\leftarrow\mathcal{F}_{SV}$.
        \For{each SV $j = 1\to N_{SV}$}
            \For{each time-step $i=1\to T$}
            \State Get the corner points of EV and SV:
            \State $\mathcal{V}_{EV}\leftarrow polygon(\{x_i,y_i,\phi_i\}_{EV},\beta)$,
            \State $\mathcal{V}_{SV}\leftarrow polygon(\{x_i,y_i,\phi_i\}_{SV},\beta)$,
            \State Collision detection based on SAT($\mathcal{V}_{EV},\mathcal{V}_{SV}$).
            \If{have collision}
                \State $C_i\leftarrow C_i+C_{init}\cdot\frac{\mathbf v_t}{\mathbf v_{base}}\cdot e^{-\mathbf{w}\beta}$
            \EndIf
            \EndFor
        \EndFor\\
        \textbf{Return:}$C_i/N_{SV}$ 
        \end{algorithmic}
        \label{A2}
    \end{algorithm}
    \end{minipage}
    \end{center}

\subsection{Comparison Baselines and Metrics}
We compare Attention embedded and Risk-aware Soft Actor Critic (ARSAC) to the following baselines:  SAC-RS \citep{ref38}, PPO-RS \citep{ref43}, which incorporate an auxiliary reward $\mathbf{r}_{safe}$ compared to standard SAC and PPO; SAC-Lag \citep{ref12} and CPO \citep{ref14}. The implementation of SAC-Lag and CPO are based on Omnisafe \citep{omnisafe}. For algorithms that do not use MMAM, the hidden layers of their policy and value networks are unified to three.  The detailed hyper-parameters of the above algorithm and the one we propose are listed in Table \ref{hyper}, with recommended values used for hyper-parameters not specified in the tables below.

\begin{table}[ht]
    \centering
    \caption{Hyper-Parameters}
    \scriptsize
    \begin{tabular}{llr}
        \toprule 
        Algorithm & Hyper-parameter & Value \\ 
        \midrule 
        \multirow{6}*{Shared} & Network hidden size &  256 \\
        ~ & Activation function & GELU \\
        ~ & Actor learning rate $\alpha_{\pi}$ & 3e-4→1e-5 \\
        ~ & (Safe) Critic learning rate $\alpha_{r,c}$ & 3e-3→1e-4 \\
        ~ & Discount factor $\gamma$ & 0.99\\
        ~ & safety threshold $d_{thres}$ & 0.05\\
        \midrule
        \multirow{5}*{SAC-RS} & Temperature factor $\tau$ &  0.005\\
        ~ & Buffer size  & 1e5\\
        ~ & Batch size  & 256\\
        ~ & Alpha learning rate $\beta_{\alpha}$ & 3e-4\\
        ~ & Target entropy $\Bar{\mathcal{H}}$ & -dim($\mathcal{A}$)\\
        \midrule
        \multirow{2}*{SAC-Lag} & Initial Lagrangian multiplier &  1.0\\
        ~ & Lagrangian multiplier learning rate $\alpha_{\lambda}$  & 1e-4\\
        \midrule
        \multirow{5}*{PPO-RS} & GAE parameter $\lambda_{GAE}$ &  0.95\\
        ~ & Clip parameter  & 0.20\\
        ~ & Batch size  & 4096\\
        ~ & Mini-batch size  & 256\\
        ~ & Activation function & TANH \\
        \midrule
        \multirow{1}*{CPO} & Conjugate gradients iterations &  15\\
        \midrule
        \multirow{3}*{ARSAC} & maximum iteration  $N_{iter}$ & 50\\
        ~ & Update step-size $\eta$ & 0.02\\
        ~ & Attention heads & 4\\
        \bottomrule
    \end{tabular}
    \label{hyper}
\end{table}

For each algorithm, five training runs with different random seeds are conducted, each spanning 10,000 episodes in randomly generated intersection scenarios featuring three driving tasks. Subsequently, each algorithm is tested with 500 episodes for LT, RT, and GS driving tasks. To evaluate the performance of EV in these intersection scenarios, this study designed the following metrics: 

 (a) \textbf{Collision rate} (CR): As the cost function devised with safety as a primary consideration, this study employs the mean collision rate as a statistical metric. Within an episode, if EV collides with other vehicles or exceeds the traversable area of the road, it is considered a collision.
 
 (b) \textbf{Success rate} (SR):  If EV safely reaches the target point without any collisions, it is considered a success. 
 
 (c) \textbf{Frozen rate} (FR): If EV neither collides nor reaches the target point within a limited time (25s), it is considered to be 'frozen'. This phenomenon is typically observed in instances where EV is operating under overly conservative strategies, which can have a detrimental impact on the overall efficiency of traffic flow. The frozen rate can be calculated by:
 \begin{equation}
\begin{aligned}[b]
FR = 1 - CR - SR.
\label{FR}
\end{aligned}
\end{equation}

(d) \textbf{Average episode cumulative reward} (AER): reflects the performance of each algorithm.
 
(e) \textbf{Average episode velocity} (AEV): reflects the average speed of the EV when navigating through intersections and its impact on traffic efficiency.

\subsection{Results Analysis}
\subsubsection{Comparison experiment}
\begin{figure}[!t]
    \centering
    \subfloat{%
        \includegraphics[width=0.30\textwidth]{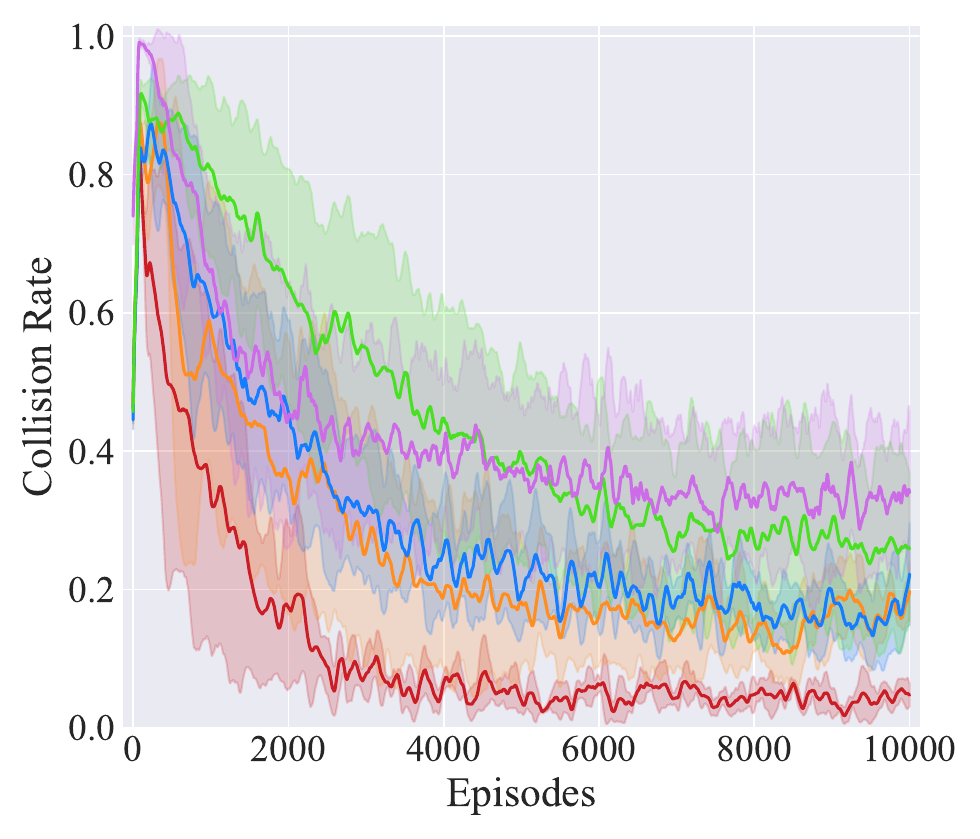}
        \label{c_c}
    }\subfloat{%
        \includegraphics[width=0.30\textwidth]{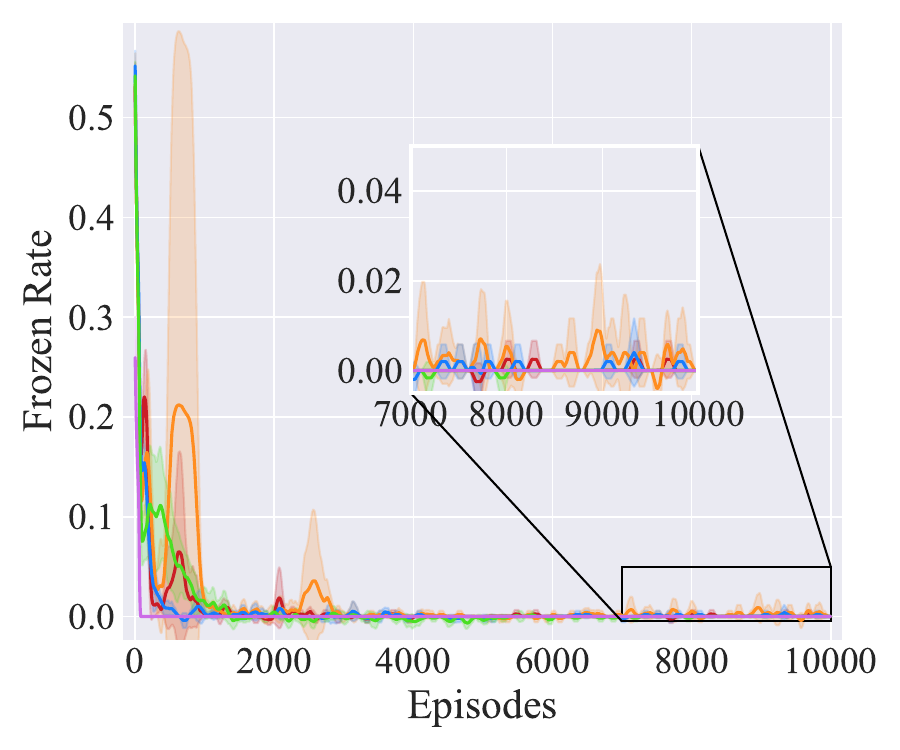}
        \label{c_f}
    }\hfill
    \subfloat{%
        \includegraphics[width=0.30\textwidth]{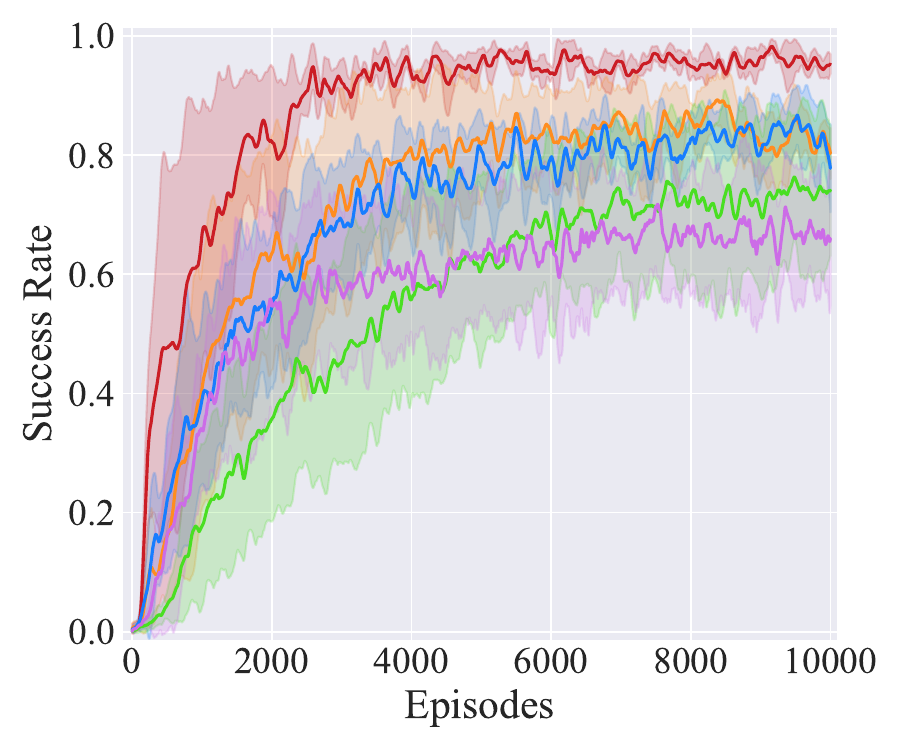}
        \label{c_s}
    }\subfloat{%
        \includegraphics[width=0.30\textwidth]{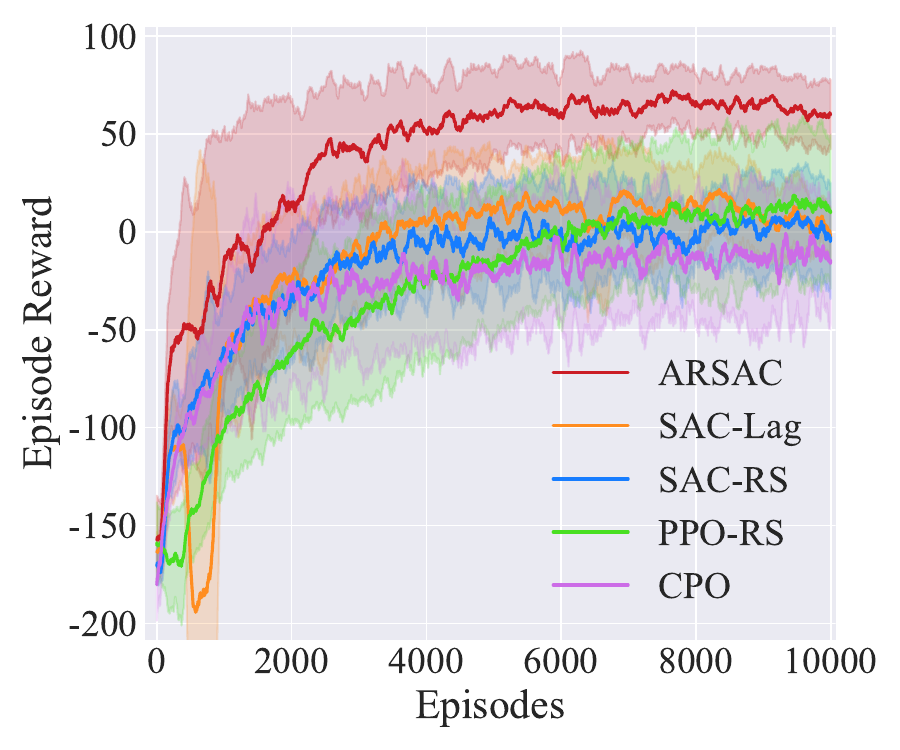}
        \label{c_r}
    }
    \caption{Training curves on comparison experiment. Solid lines correspond to the mean and the shaded regions correspond to 95\% confidence interval over 3 runs.}
    \label{f_compare}
\end{figure}
The learning curves compared with baseline algorithms such as SAC-RS, PPO-RS, SAC-Lag and CPO are shown in Fig. \ref{f_compare} and test results are in Table \ref{t_compare}. Results indicate that the proposed ARSAC algorithm outperforms or matches all other baseline algorithms across three driving tasks in terms of the final performance. For instance, in the RT task, ARSAC achieves 18.2\%, 16.4\%, 10.2\%, and 8.4\% lower CR compared to CPO, PPO-RS, SAC-RS, and SAC-Lag respectively, while demonstrating superior performance in AER with significantly lower variance. In the LT task, SAC-RS, PPO-RS, SAC-Lag and CPO demonstrate higher velocity, yet their AER are negative and significantly lower than that of ARSAC. In this scenario, EV needs to adopt a competitive strategy to efficiently identify suitable gaps in fast-moving traffic. Due to insufficient understanding of the scenario, SAC-RS, PPO-RS, SAC-Lag and CPO struggle to achieve higher rewards, resulting in mostly negative outcomes. To maximize cumulative rewards, these algorithms tend to adopt higher driving speeds to avoid accumulating negative rewards in future timesteps, leading to higher AEV and undesirable behaviors such as failure to reach the destination. Although CPO and PPO-RS achieve or surpass ARSAC's performance in FR in both the LT and GS tasks, they lack safety and perform worse than ARSAC in terms of AER. Additionally, in Table \ref{t_compare}, we present the mean statistics that evaluate the average performance of each method across three testing conditions. We find that ARSAC outperforms or matches the baselines across the three tasks and demonstrates exceptional performance in the more challenging unprotected left-turn scenario.

\begin{table}
    \setlength\tabcolsep{12pt}
    \centering
    \caption{Compare Performance on Three Driving Tasks.}
    \begin{threeparttable}
    \scriptsize
    \begin{tabular}{c|c|ccccc}
        \toprule 
        Tasks & Algorithms & CR(\%) 
 & SR(\%) & FR(\%) & AER & AEV(m/s) \\ 
        \midrule 
        \multirow{5}*{LT} & CPO & 38.4$\pm$10.8 & 61.6$\pm$10.8 & \textbf{0.0$\pm$0.0} & -91.2$\pm$48.3 & \underline{14.12}$\pm$0.82\\
        ~ & PPO-RS & 31.4$\pm$6.6 & 68.6$\pm$6.6 & \textbf{0.0$\pm$0.0} & -78.3$\pm$38.3 & \underline{13.56}$\pm$0.48\\
        ~ & SAC-RS & 17.6$\pm$5.1 & 80.8$\pm$5.1 & 1.6$\pm$0.3 & -49.4$\pm$30.1 & \underline{12.69}$\pm$0.55 \\
        ~ & SAC-Lag & 17.4$\pm$4.5 & 80.4$\pm$4.4 & 2.2$\pm$1.1 & -37.1$\pm$32.4 & \underline{12.16}$\pm$0.49\\
        ~ & ARSAC & \textbf{2.8$\pm$1.7} & \textbf{95.8$\pm$1.3} & 1.4$\pm$0.5 & 46.32$\pm$19.9 & \textbf{8.51$\pm$0.76}\\
        \midrule 
        \multirow{5}*{GS} & CPO & 20.8$\pm$8.1 & 79.2$\pm$8.1 & \textbf{0.0$\pm$0.0} & 41.32$\pm$31.6 & \textbf{8.82$\pm$1.62}\\
        ~ & PPO-RS & 14.4$\pm$6.0 & 85.6$\pm$6.0 & \textbf{0.0$\pm$0.0} & 54.80$\pm$35.9 & 7.82$\pm$0.32\\
        ~ & SAC-RS & 12.8$\pm$3.0 & 86.2$\pm$3.1 & 1.0$\pm$0.6 & 46.53$\pm$23.9 & 7.54$\pm$0.88 \\
        ~ & SAC-Lag & 11.6$\pm$4.4 & 87.2$\pm$4.0 & 1.2$\pm$0.8 & 48.42$\pm$30.1 & 8.01$\pm$0.46\\
        ~ & ARSAC & \textbf{1.6$\pm$1.0} & \textbf{98.4$\pm$1.0} & \textbf{0.0$\pm$0.0} & \textbf{76.62$\pm$17.4} & 8.23$\pm$0.28\\
        \midrule 
        \multirow{5}*{RT} & CPO & 18.6$\pm$7.1 & 81.0$\pm$7.1 & 0.4$\pm$0.1 & 41.5$\pm$37.2 & \textbf{9.10$\pm$0.86}\\
        ~ & PPO-RS & 16.8$\pm$5.7 & 82.4$\pm$5.6 & 0.8$\pm$0.1 & 58.5$\pm$35.8 & 8.32$\pm$0.47\\
        ~ & SAC-RS & 10.6$\pm$3.7 & 87.8$\pm$3.7 & 1.6$\pm$0.6 & 52.1$\pm$26.5 & 8.46$\pm$0.55\\
        ~ & SAC-Lag & 8.8$\pm$2.8 & 89.6$\pm$2.8 & 1.6$\pm$0.3 & 49.9$\pm$22.6 & 7.68$\pm$1.13\\
        ~ & ARSAC & \textbf{0.4$\pm$0.3} & \textbf{99.6$\pm$0.3} & \textbf{0.0$\pm$0.0} & \textbf{90.4$\pm$12.4} & 8.27$\pm$0.35\\
        \midrule 
        \multirow{5}*{MEAN} & CPO & 25.9$\pm$8.8 & 73.9$\pm$8.8 & \textbf{0.1$\pm$0.0} & -2.8$\pm$41.2 & \textbf{10.68$\pm$1.02}\\
        ~ & PPO-RS & 20.9$\pm$6.2 & 78.9$\pm$6.1 & 0.3$\pm$0.0 & 11.67$\pm$36.9 & 9.90$\pm$0.39\\
        ~ & SAC-RS & 13.7$\pm$4.2 & 84.9$\pm$4.0 & 1.4$\pm$0.4 & 16.43$\pm$27.1 & 9.56$\pm$0.68\\
        ~ & SAC-Lag & 12.6$\pm$4.1 & 85.7$\pm$4.2 & 1.7$\pm$0.7 & 20.41$\pm$24.8 & 9.28$\pm$0.89\\
        ~ & ARSAC & \textbf{1.6$\pm$0.9} & \textbf{97.9$\pm$0.9} & 0.5$\pm$0.1 & \textbf{71.13$\pm$16.8} & 8.37$\pm$0.45\\
        \bottomrule
    \end{tabular}
    \begin{tablenotes}
    \footnotesize
    \item[1] Bold: best performance; Underline: undesirable high values. 
    \item[2] Policy update frequency $f_{\pi} = 10$ Hz.
    \end{tablenotes}
    \end{threeparttable}
    \label{t_compare}
\end{table}

\subsubsection{Ablation studies}

\begin{figure}[!t]
    \centering
    \subfloat{%
        \includegraphics[width=0.30\textwidth]{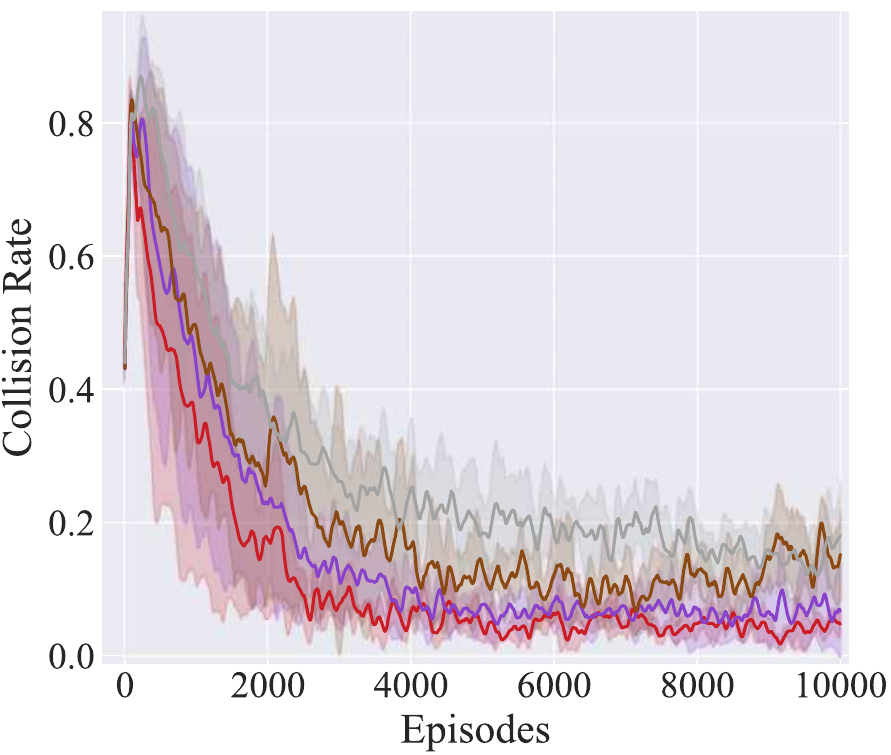}
        \label{a_c}
    }\subfloat{%
        \includegraphics[width=0.30\textwidth]{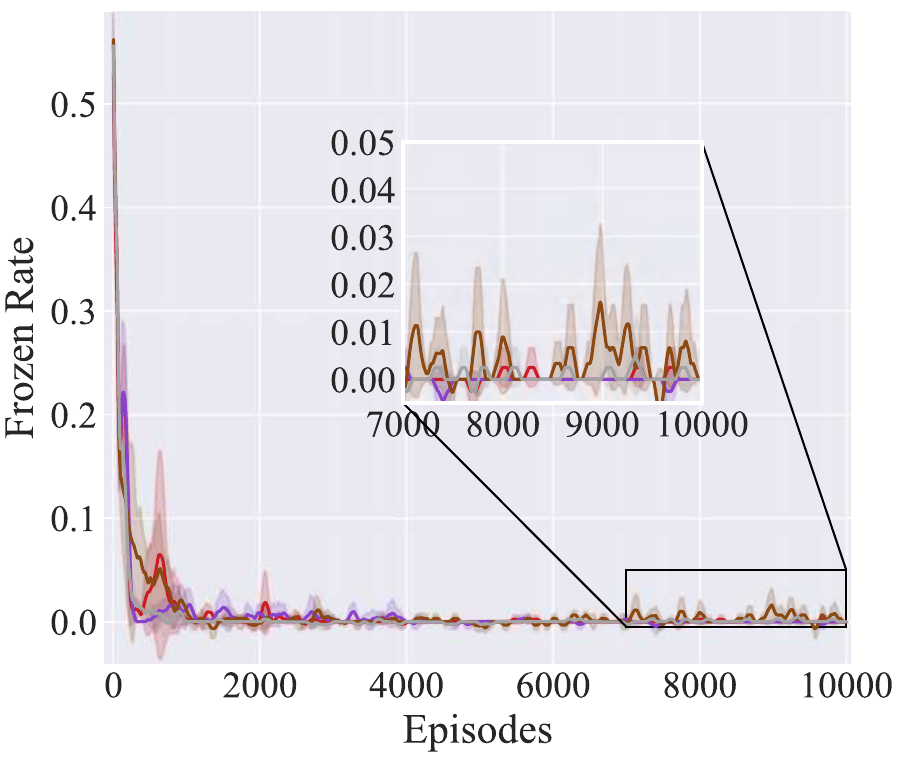}
        \label{a_f}
    }\hfill
    \subfloat{%
        \includegraphics[width=0.30\textwidth]{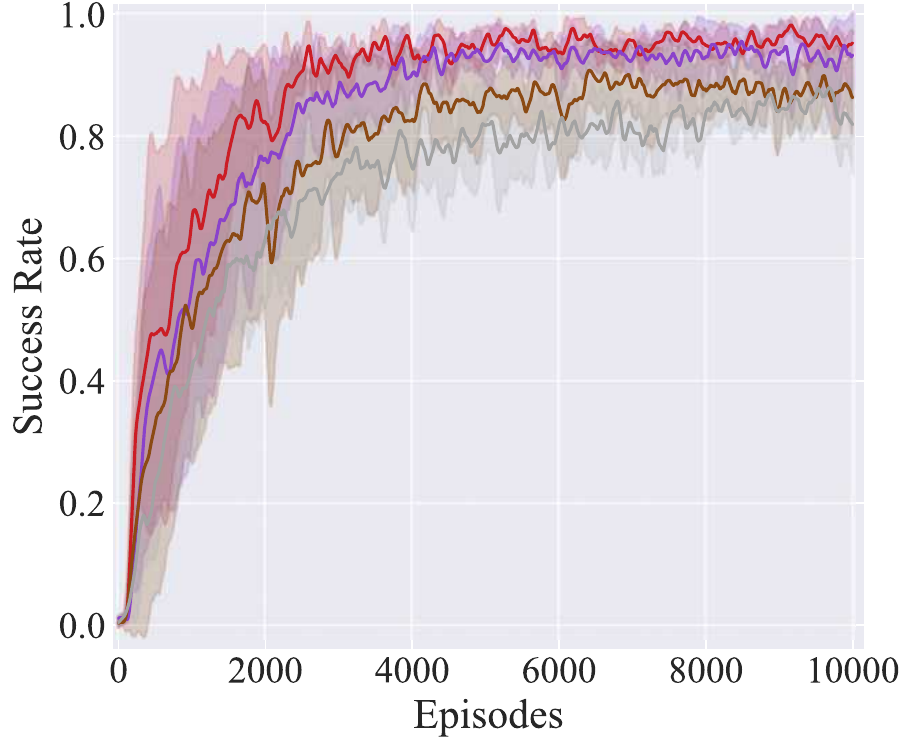}
        \label{a_s}
    }\subfloat{%
        \includegraphics[width=0.30\textwidth]{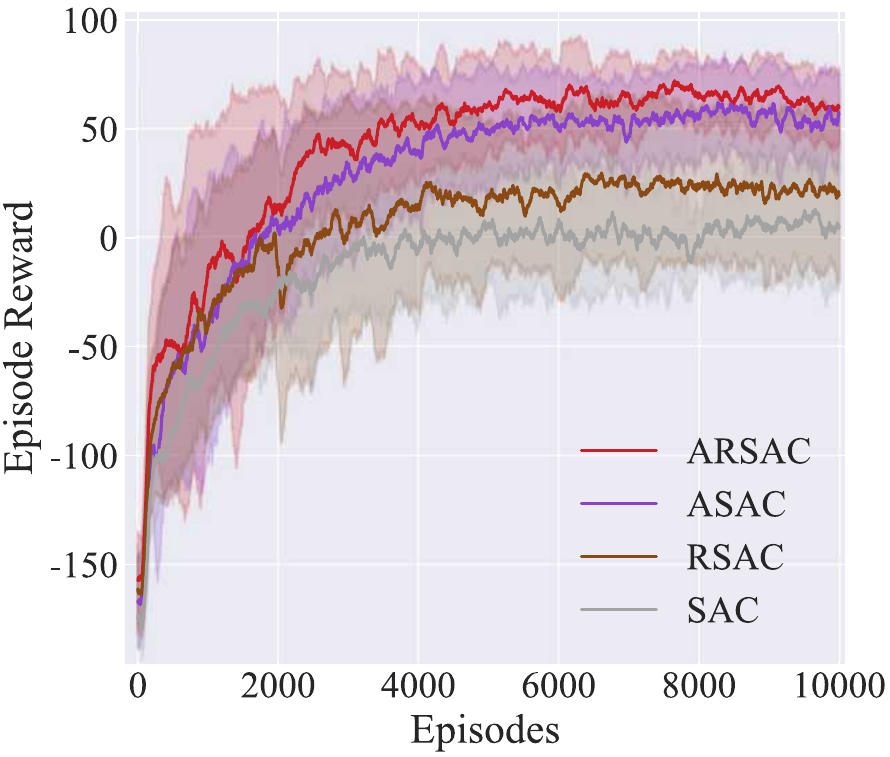}
        \label{a_r}
    }
    \caption{Training curves on ablation studies. Solid lines correspond to the mean and the shaded regions correspond to 95\% confidence interval over 3 runs.}
    \label{f_ablation}
\end{figure}

\begin{table}[ht]
    \setlength\tabcolsep{12pt}
    \centering
    \caption{Ablation Study on Three Driving Tasks.}
    \begin{threeparttable}
    \scriptsize
    \begin{tabular}{c|c|ccccc}
        \toprule 
        Tasks & Algorithms & CR(\%) 
 & SR(\%) & FR(\%) & AER & AEV(m/s) \\ 
        \midrule 
        \multirow{4}*{LT} & SAC & 20.4$\pm$5.4 & 79.2$\pm$5.3 & 0.4$\pm$0.0 & -42.24$\pm$31.2 & \underline{12.28}$\pm$0.69 \\
        ~ & RSAC & 13.2$\pm$4.8 & 84.2$\pm$4.6 & 2.6$\pm$0.3 & -34.28$\pm$27.8 & \underline{10.96}$\pm$0.56 \\
        ~ & ASAC & 4.8$\pm$1.2 & 95.2$\pm$1.2 & \textbf{0.0$\pm$0.0} & 41.43$\pm$24.9 & \textbf{8.63$\pm$0.68} \\
        ~ & ARSAC & \textbf{2.8$\pm$1.7} & \textbf{95.8$\pm$1.3} & 1.4$\pm$0.5 & 46.32$\pm$19.9 & 8.51$\pm$0.76\\
        \midrule 
        \multirow{4}*{GS} & SAC & 14.6$\pm$5.9 & 83.2$\pm$5.5 & 2.2$\pm$0.4 & 51.02$\pm$34.3 & 7.95$\pm$0.46 \\
         ~ & RSAC & 5.4$\pm$1.6 & 93.2$\pm$1.2 & 1.4$\pm$0.3 & 66.74$\pm$29.1 & 7.69$\pm$0.37 \\
        ~ & ASAC & 4.6$\pm$0.9 & 95.4$\pm$0.9 & \textbf{0.0$\pm$0.0} & 69.42$\pm$23.6 & \textbf{8.42$\pm$0.33} \\
        ~ & ARSAC & \textbf{1.6$\pm$1.0} & \textbf{98.4$\pm$1.0} & \textbf{0.0$\pm$0.0} & \textbf{76.62$\pm$17.4} & 8.23$\pm$0.28\\
        \midrule 
        \multirow{4}*{RT} & SAC & 12.6$\pm$4.8 & 85.6$\pm$4.8 & 1.8$\pm$0.3 & 58.59$\pm$31.7 & 7.84$\pm$0.46\\
        ~ & RSAC & 4.2$\pm$1.4 & 94.6$\pm$1.3 & 1.2$\pm$0.1 & 76.84$\pm$24.2 & 8.13$\pm$0.33\\
        ~ & ASAC & 3.6$\pm$0.7 & 96.4$\pm$0.7 & \textbf{0.0$\pm$0.0} & 86.85$\pm$13.2 & \textbf{8.50$\pm$0.61} \\
        ~ & ARSAC & \textbf{0.4$\pm$0.3} & \textbf{99.6$\pm$0.3} & \textbf{0.0$\pm$0.0} & \textbf{90.4$\pm$12.4} & 8.27$\pm$0.35\\
        \midrule 
        \multirow{4}*{MEAN} & SAC & 15.2$\pm$5.2 & 83.3$\pm$5.1 & 1.5$\pm$0.3 & 22.46$\pm$32.2 & 9.36$\pm$0.52\\
        ~ & RSAC & 7.6$\pm$3.3 & 90.7$\pm$3.1 & 1.7$\pm$0.5 & 36.43$\pm$25.2 & 8.93$\pm$0.41\\
        ~ & ASAC & 4.3$\pm$0.8 & 95.7$\pm$0.8 & \textbf{0.0$\pm$0.0} & 65.90$\pm$19.2 & \textbf{8.52}$\pm$0.49 \\
        ~ & ARSAC & \textbf{1.6$\pm$0.9} & \textbf{97.9$\pm$0.9} & 0.5$\pm$0.1 & \textbf{71.13$\pm$16.8} & 8.37$\pm$0.45\\
        \bottomrule
    \end{tabular}
    \begin{tablenotes}
    \footnotesize
    \item[1] Bold: best performance; Underline: undesirable high values. 
    \item[2] Policy update frequency $f_{\pi} = 10$ Hz.
    \end{tablenotes}
    \end{threeparttable}
    \label{t_ablation}
\end{table}

We additionally perform an ablation study to compare the effects of the safety module and the structure of MMAM on algorithm performance. As shown in Table \ref{t_ablation}, RSAC is a variant of ARSAC that excludes the MMAM, while ASAC is a version of ARSAC that omits the risk-aware component.  Compared to SAC, RSAC demonstrates a lower collision rate across three driving tasks. Although RSAC encounters performance degradation in LT task due to limited scene comprehension. Safety iterative correction mitigates collisions by projecting risky actions back towards the feasible region $\mathcal{S}_{FR}$ through cyclic gradient descent, guided by safe critics' evaluations. The collision rates in LT/GS/RT task are reduced by 2.0\%, 3.0\%, and 3.2\% for ARSAC  compared to ASAC, while the collision rates are reduced by 10.4\%, 3.8\%, and 3.8\% in comparison to RSAC, respectively. These results indicate that better scene understanding allows the safe critic to more accurately assess risky actions. Consequently, the safe actions corrected by gradient projection are more likely to fall within $\mathcal{S}_{FR}$, thereby enhancing overall safety. As illustrated in Fig. \ref{f_ablation}, after training convergence,the frozen rate of RSAC exhibits fluctuations around 1\%, while ARSAC exhibits a notable decline in its frozen rate.  Furthermore, as indicated in Table \ref{t_ablation}, the AEV of RSAC is 1.7\% and 6.5\% lower than that of ARSAC in RT and GS tasks, respectively. In contrast, ASAC achieves AEV that are 2.8\% and 2.3\% higher than those of ARSAC. These findings illustrate that MMAM effectively captures the relationship between EV and SVs, filtering out non-conflicting vehicles and other disruptive factors, thereby enhancing overall traffic efficiency. 

\begin{figure*}[!t]
    \centering
    \subfloat[]{%
        \includegraphics[width=0.19\textwidth]{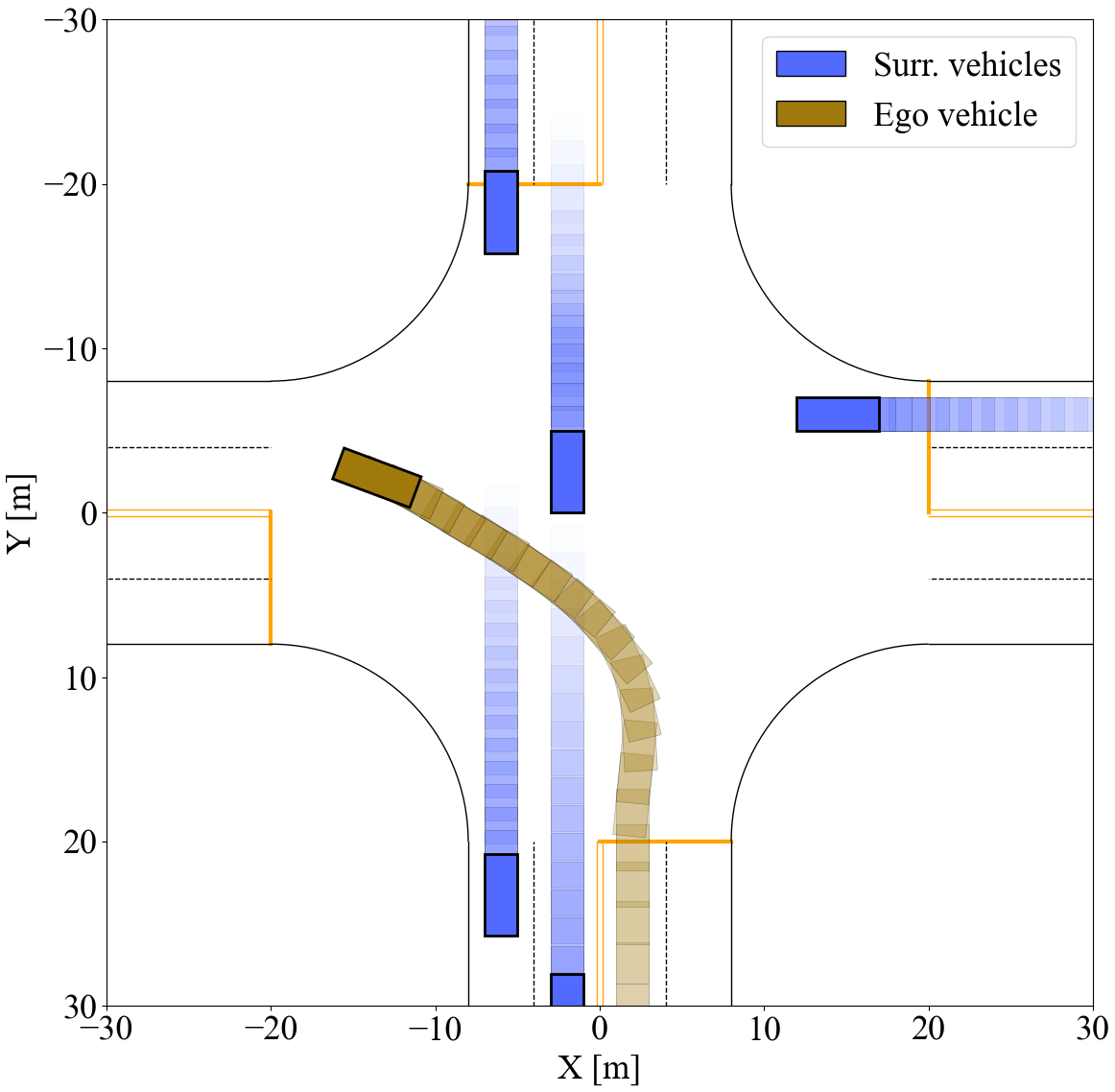}
        \label{left_arsac}
    }\subfloat[]{%
        \includegraphics[width=0.19\textwidth]{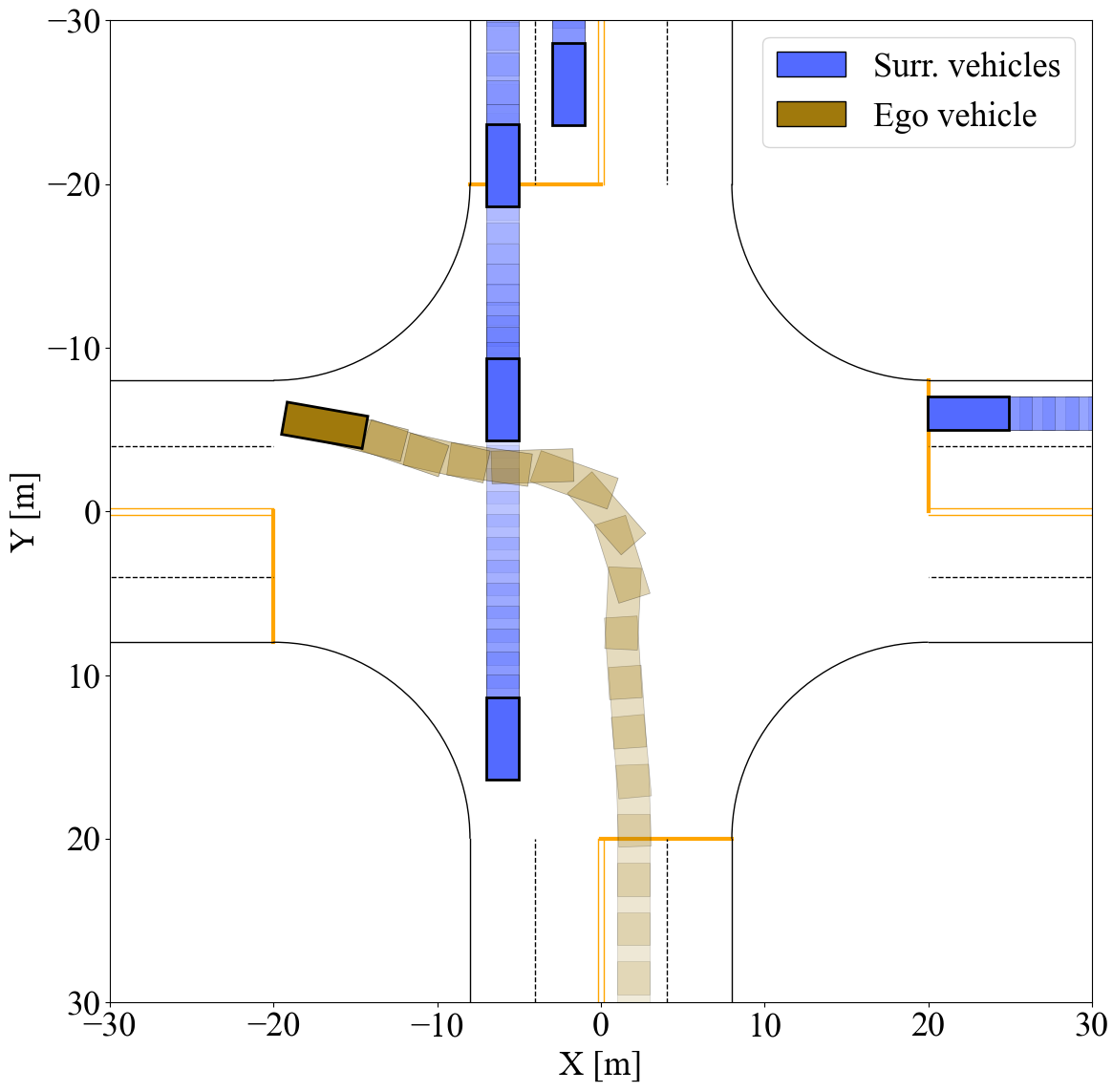}
        \label{left_saclag}
    }\subfloat[]{%
        \includegraphics[width=0.19\textwidth]{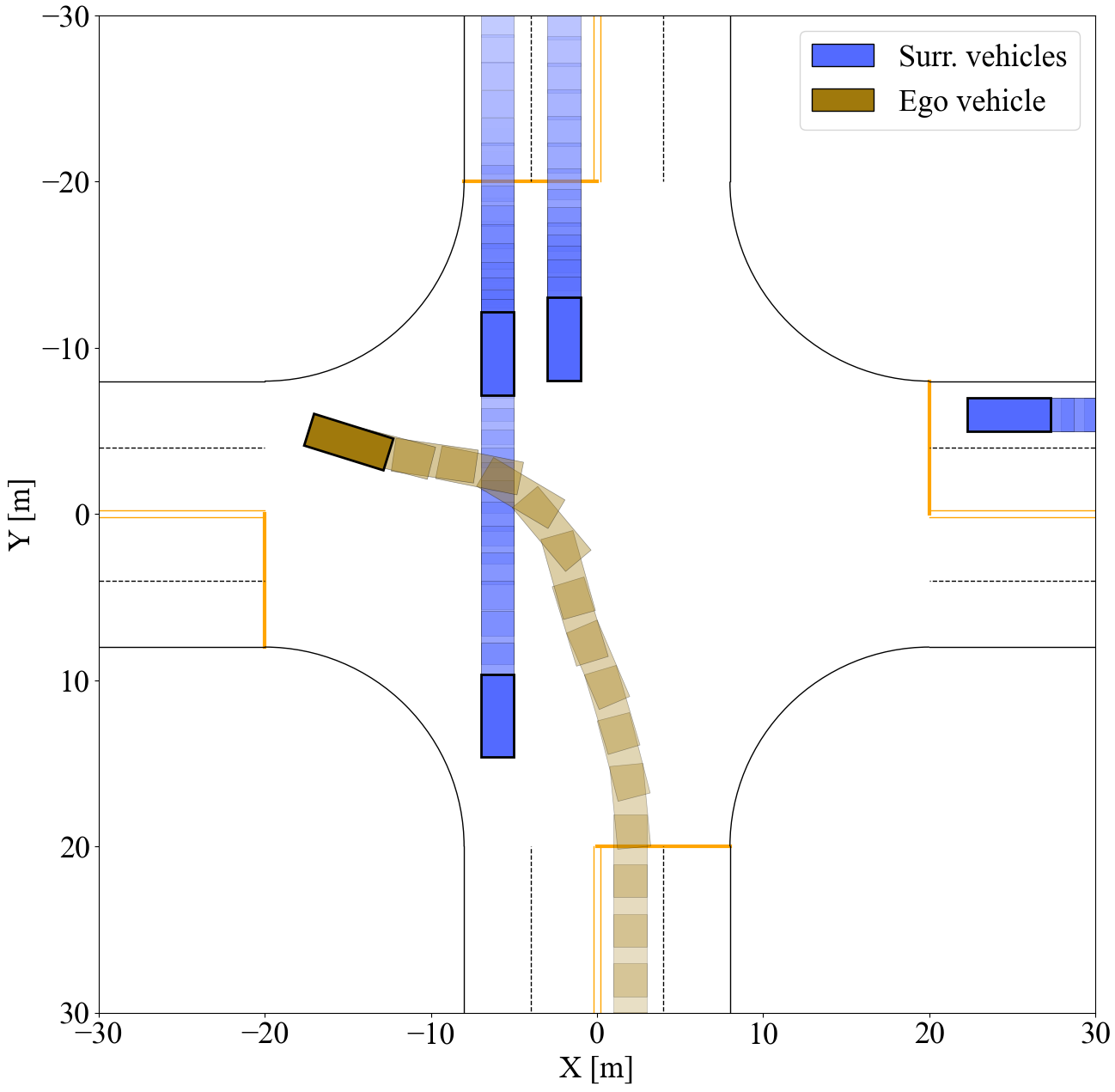}
        \label{left_sacrs}
    }\subfloat[]{%
        \includegraphics[width=0.19\textwidth]{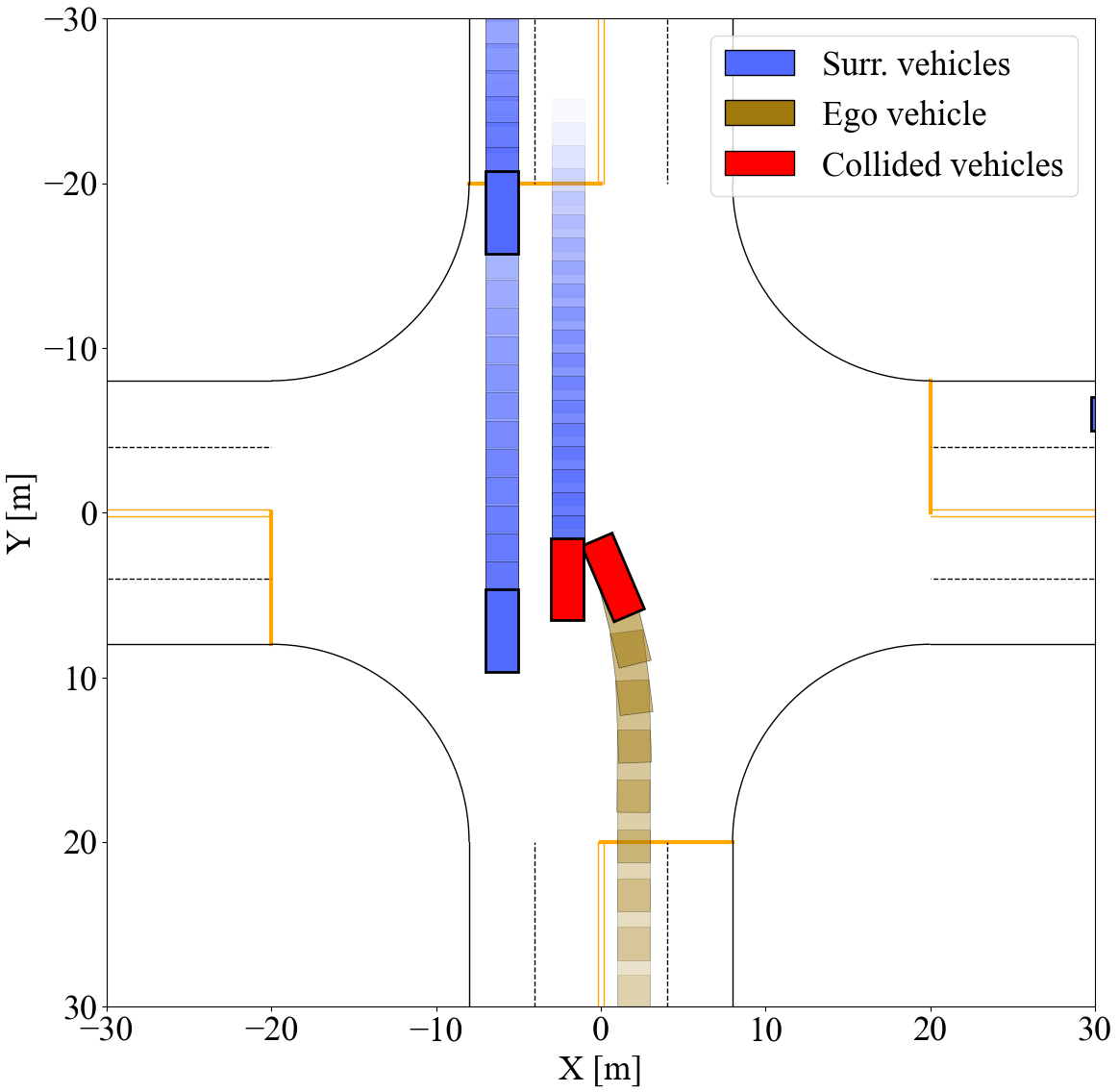}
        \label{left_ppors}
    }\subfloat[]{%
        \includegraphics[width=0.19\textwidth]{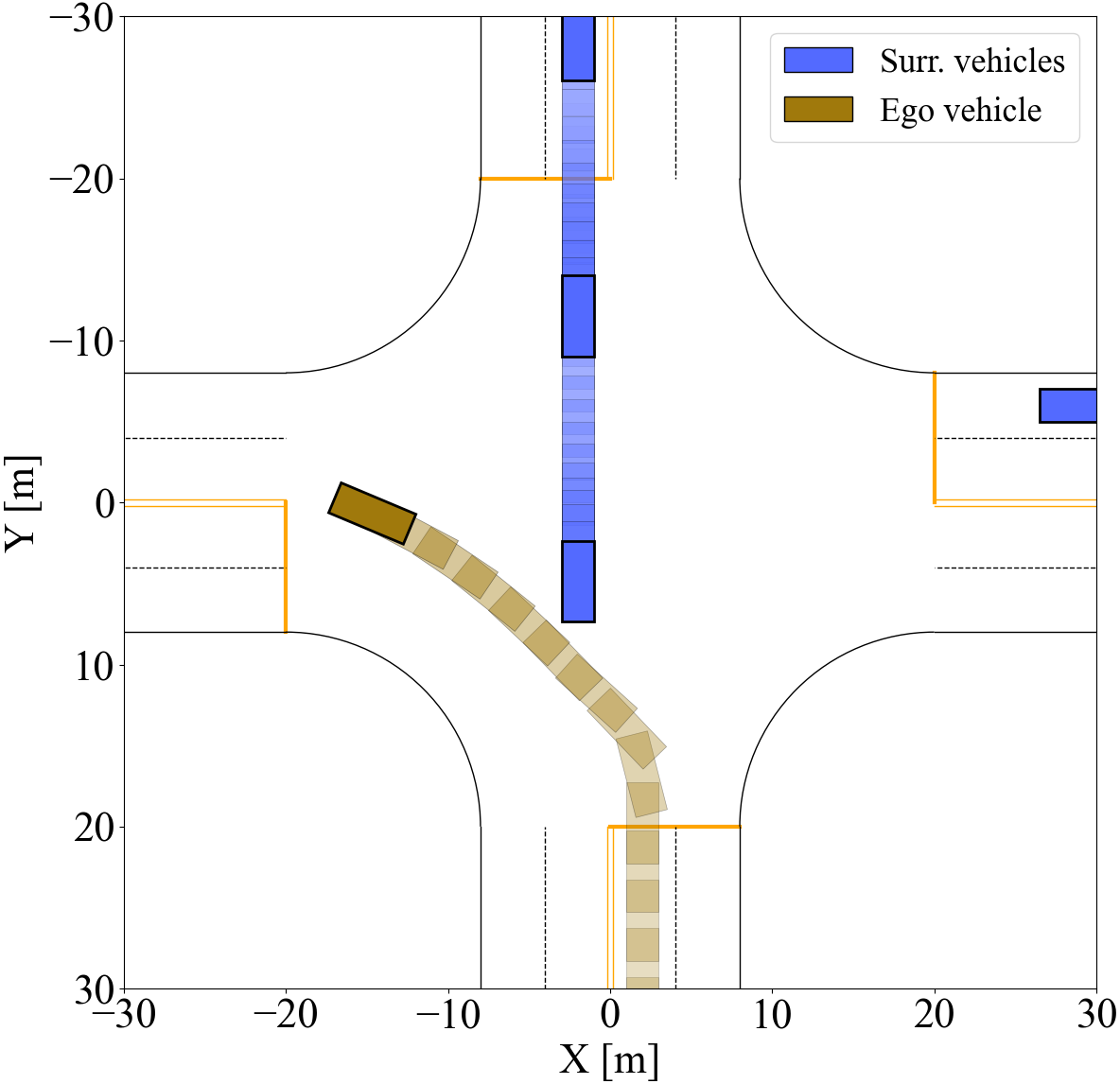}
        \label{left_cpo}
    }\\ \vspace{-5pt}
    \subfloat[]{%
        \includegraphics[width=0.98\textwidth]{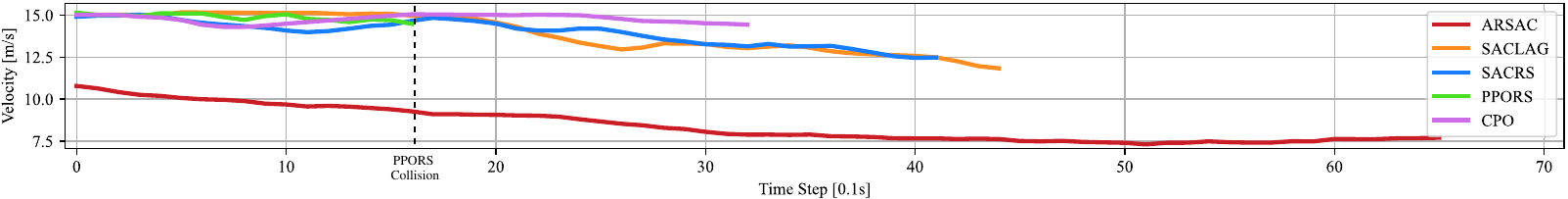}
        \label{left_velocity}
    }
    \caption{Vehicle trajectories visualization. Blue rectangles are SVs, brown rectangles are EVs. (a) ARSAC. (b) SAC-Lag. (c) SAC-RS. (d) PPO-RS. (e) CPO. (f) The velocity of the EV driving inside intersection.}
    \label{slice_compare_left}
\end{figure*}

\subsection{Driving Behavior Analysis}
We apply the trained policies of the compared baselines and our proposed algorithm, ARSAC, to three driving tasks and visualize the trajectories of both EVs and SVs. The basic environment settings are consistent with those described in Sec. \ref{subsec:env_set}.
\subsubsection{Left-turn Case}
 As illustrated in Fig. \ref{slice_compare_left}, ARSAC, SAC-Lag, SAC-RS and CPO are able to pass through the scene without collisions, while PPO-RS encounters a collision. ARSAC exhibits similar driving behaviour to a human driver when faced with oncoming traffic in parallel lanes. When approaching an intersection, ARSAC first slows down and then performs a pre-steer maneuver to the right, allowing the vehicle to turn left more fluidly and safely. During the turn, it decelerates appropriately to find an optimal moment to pass through, and once the oncoming traffic is cleared, it accelerates again to improve traffic flow. Throughout the turn, the ego's trajectory follows a smooth arc. In contrast, the trajectory of SAC-RS is not smooth, and although SAC-Lag also exhibits a tendency to pre-steer rightward, its trajectory is similarly not smooth. Due to its high-speed performance, CPO maneuvers left to avoid oncoming vehicles. However, in comparison to ARSAC, its trajectory deviates from the reference line. It can be seen that ARSAC enables the vehicle to effectively perceive its surroundings, thereby enhancing safety and providing improved opportunities for better passage.

\subsubsection{Go-straight Case}
\begin{figure*}[!t]
    \centering
    \subfloat[]{%
        \includegraphics[width=0.19\textwidth]{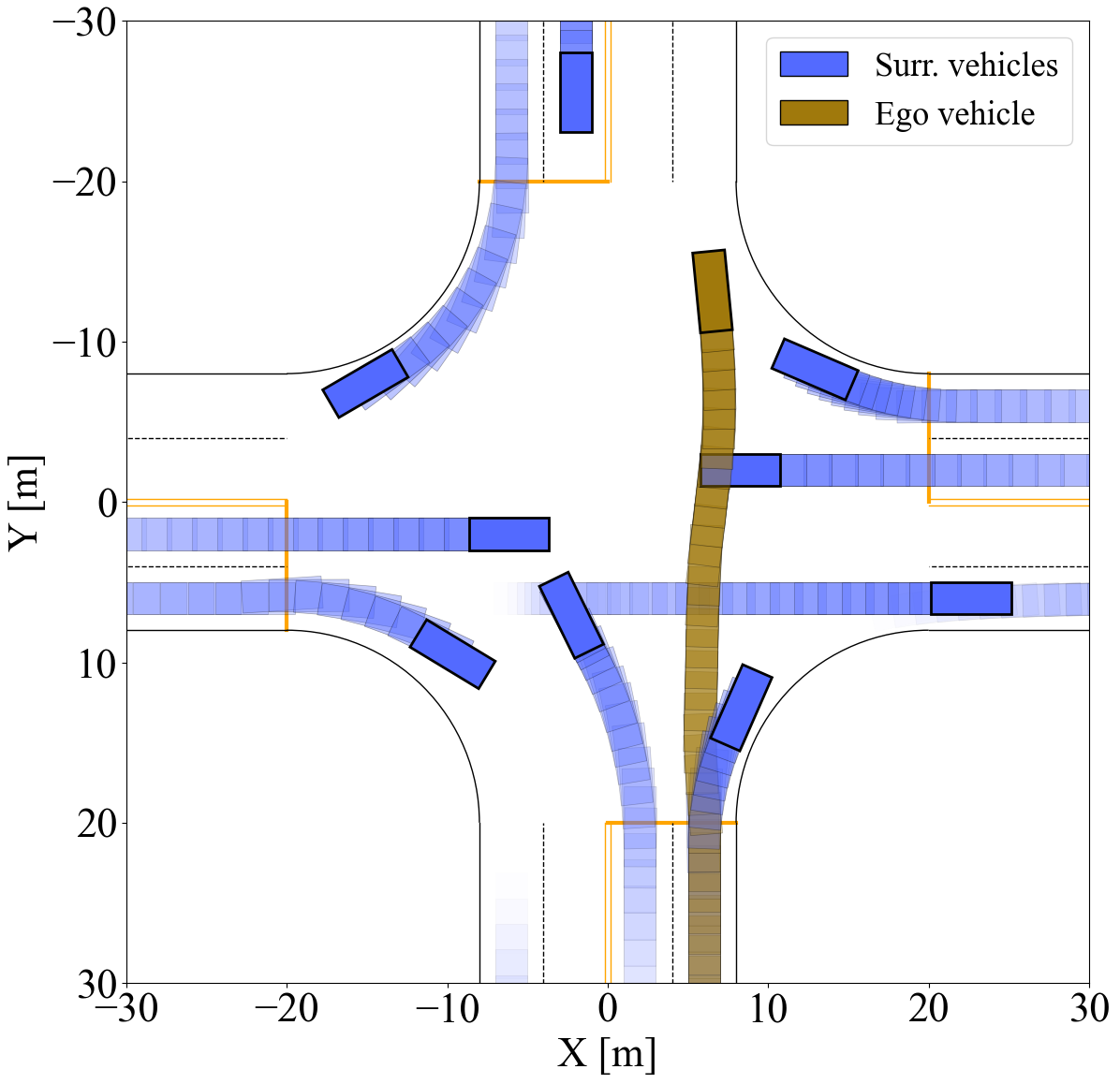}
        \label{straight_arsac}
    }\subfloat[]{%
        \includegraphics[width=0.19\textwidth]{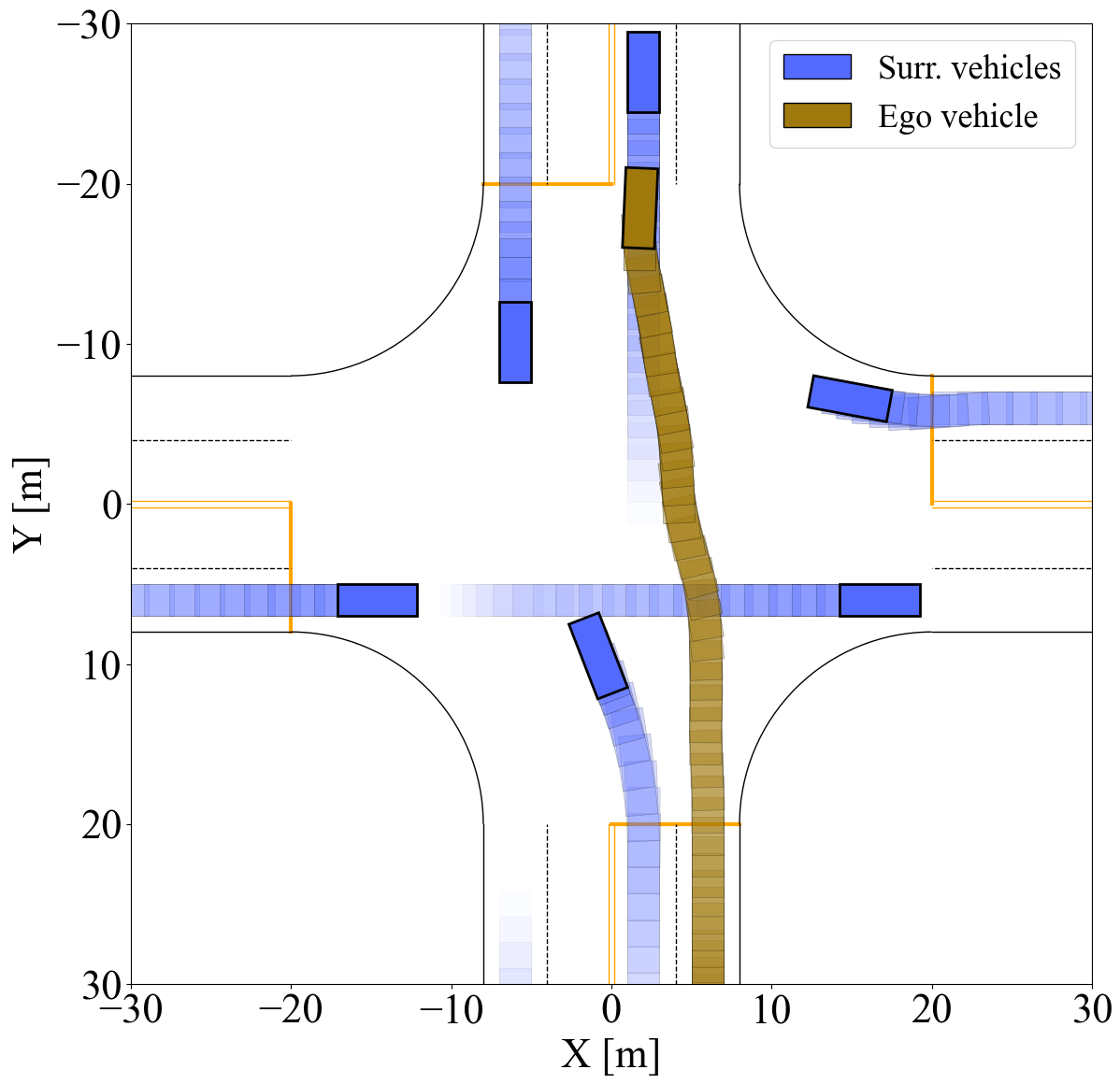}
        \label{straight_saclag}
    }\subfloat[]{%
        \includegraphics[width=0.19\textwidth]{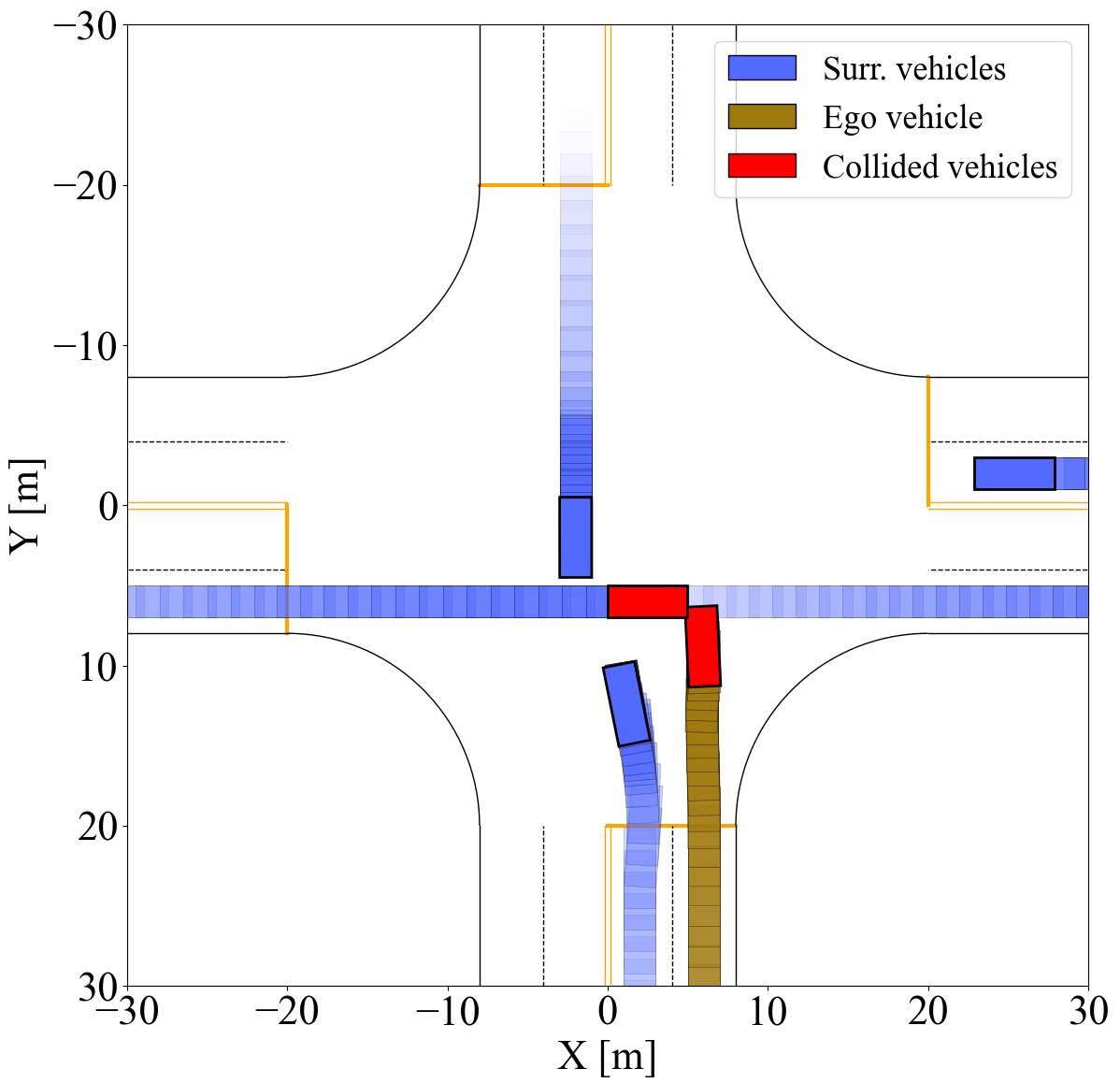}
        \label{straight_sacrs}
    }\subfloat[]{%
        \includegraphics[width=0.19\textwidth]{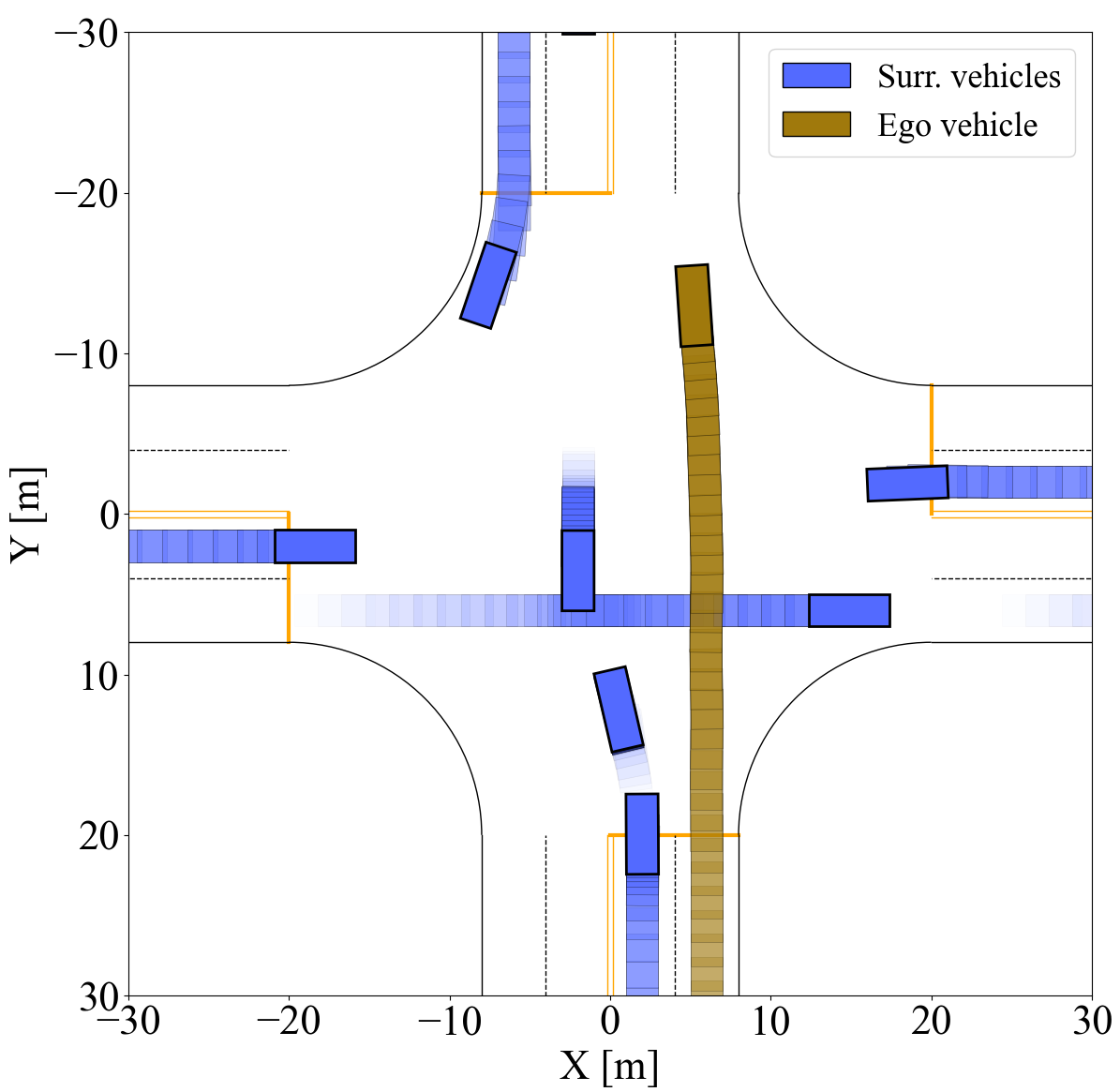}
        \label{straight_ppors}
    }\subfloat[]{%
        \includegraphics[width=0.19\textwidth]{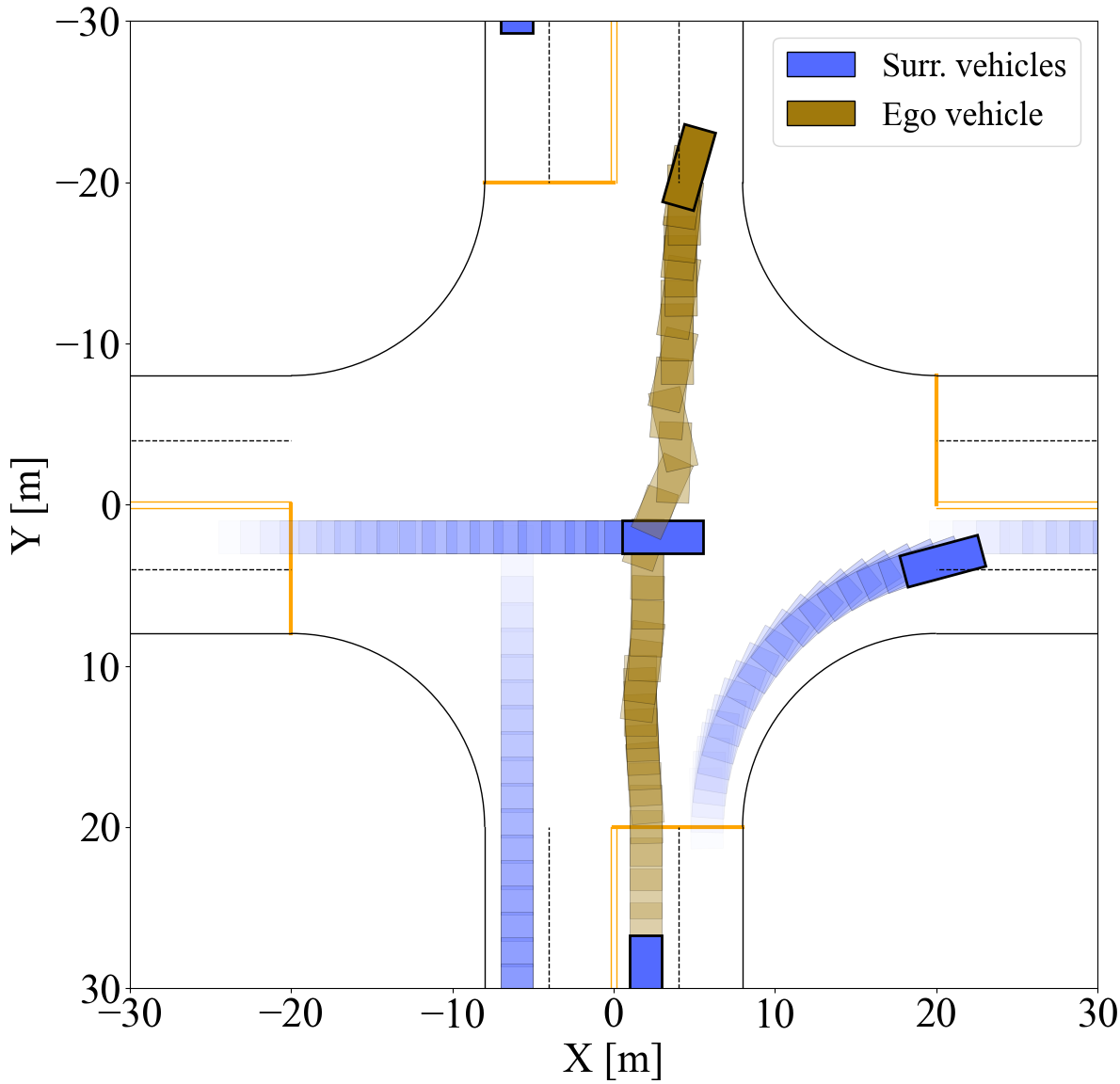}
        \label{straight_cpo}
    }\\ \vspace{-5pt}
    \subfloat[]{%
        \includegraphics[width=0.98\textwidth]{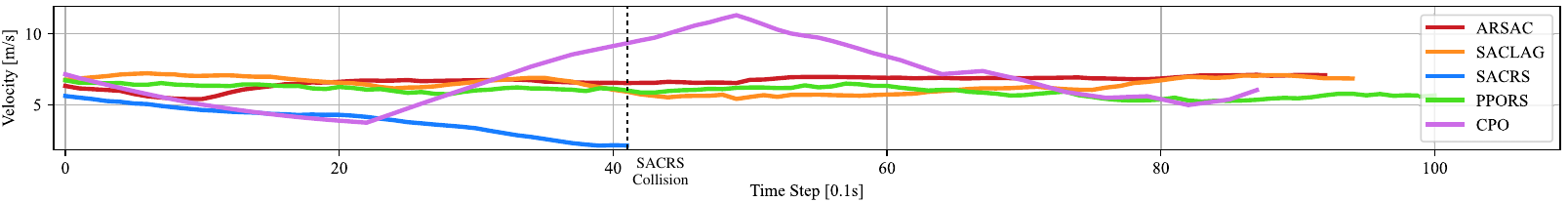}
        \label{straight_velocity}
    }
    \caption{Vehicle trajectories visualization. Blue rectangles are SVs, brown rectangles are EVs. (a) ARSAC. (b) SAC-Lag. (c) SAC-RS. (d) PPO-RS. (e) CPO. (f) The velocity of the EV driving inside intersection.}
    \label{slice_compare_straight}
\end{figure*}

In the GS task, the EV encounters the challenge of traffic coming from all directions. If the EV fails to respond expeditiously to potential risks in its surroundings, the likelihood of a collision occurring is increased. As shown in Fig. \subref*{straight_sacrs}, Although SAC-RS attempts to avoid collisions with oncoming lateral traffic by reducing speed, the utilization of $\mathbf{r}_{safe}$ as an auxiliary reward alone is inadequate to ensure collision avoidance. In this case, CPO made significant accelerations and decelerations to avoid a collision. Although it successfully navigated through the intersection, its trajectory was less smooth compared to ARSAC, and the larger speed fluctuations could result in a decrease in comfort. Fig. \subref*{straight_velocity} demonstrates that the velocity changes of ARSAC, SAC-Lag, and PPO-RS are characterized by a relatively gentle slope. Nevertheless, due to its limited scene understanding, SAC-Lag is compelled to execute preemptive steering maneuvers to avoid collisions with oncoming lateral traffic, causing the EV's trajectory to shift left. In contrast, ARSAC effectively seizes the opportunity to pass through, making minimal adjustments to avoid collisions while maintaining a smooth trajectory.

\begin{figure*}[!t]
    \centering
    \subfloat[]{%
        \includegraphics[width=0.19\textwidth]{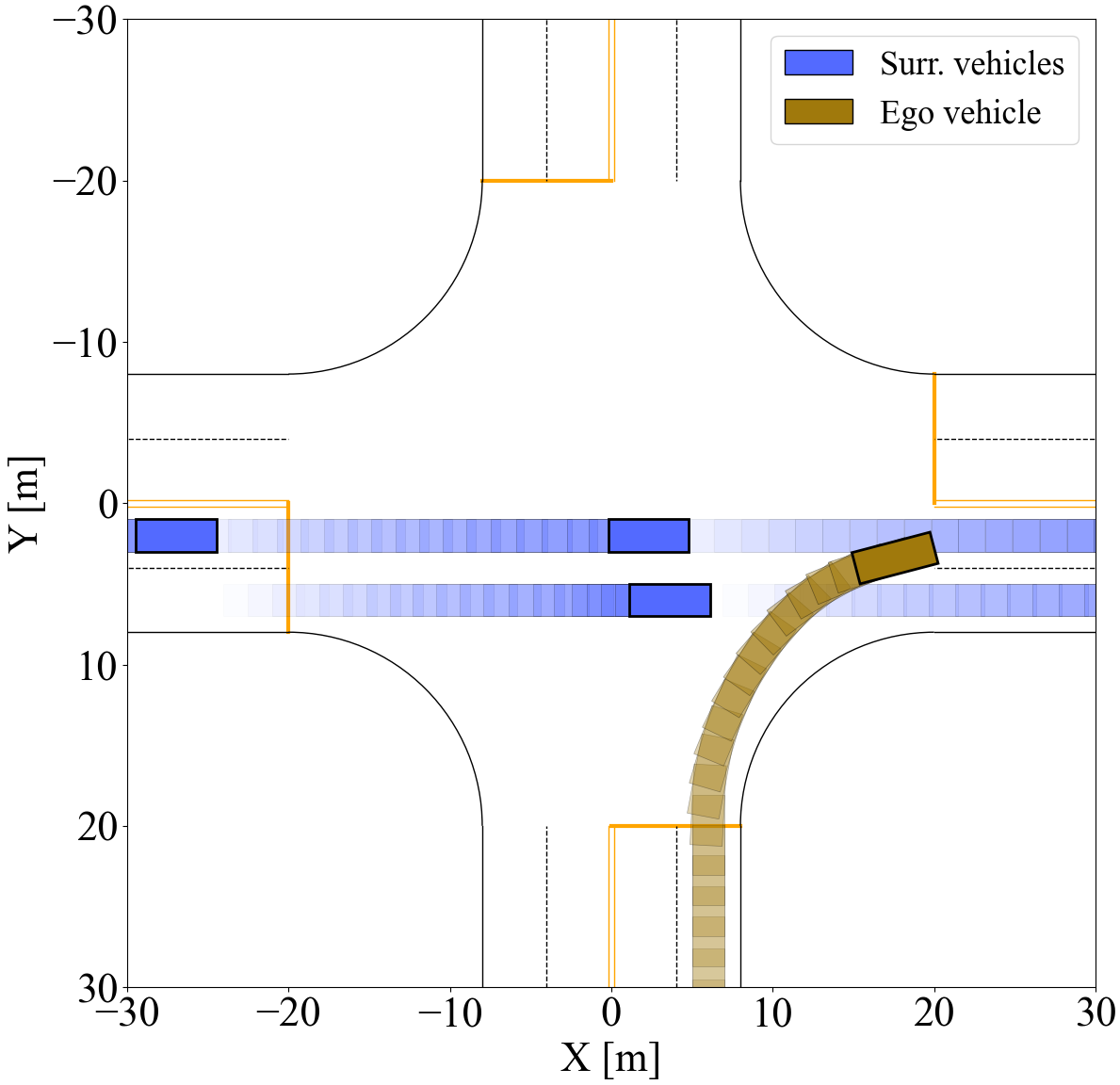}
        \label{right_arsac}
    }\subfloat[]{%
        \includegraphics[width=0.19\textwidth]{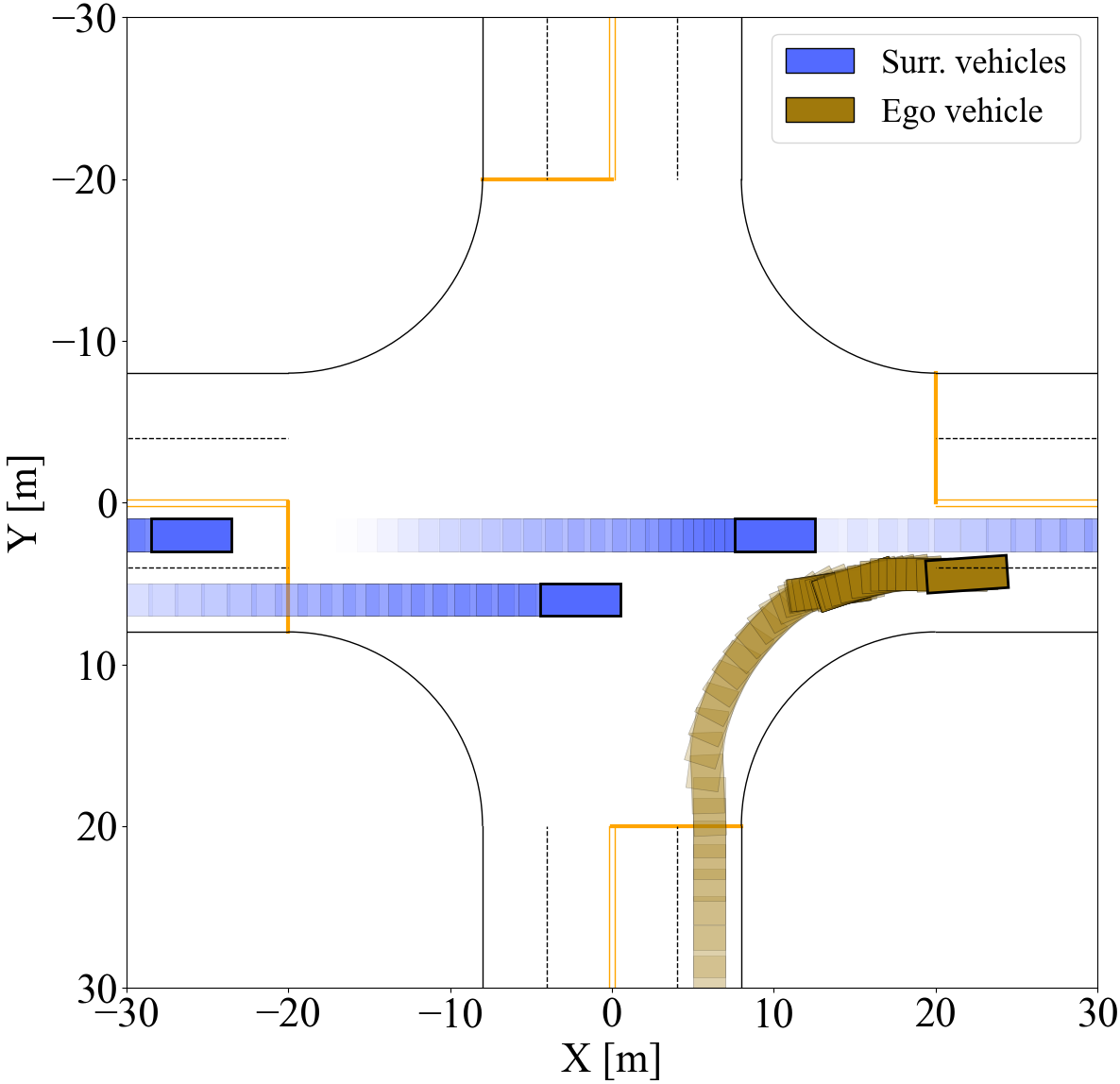}
        \label{right_saclag}
    }\subfloat[]{%
        \includegraphics[width=0.19\textwidth]{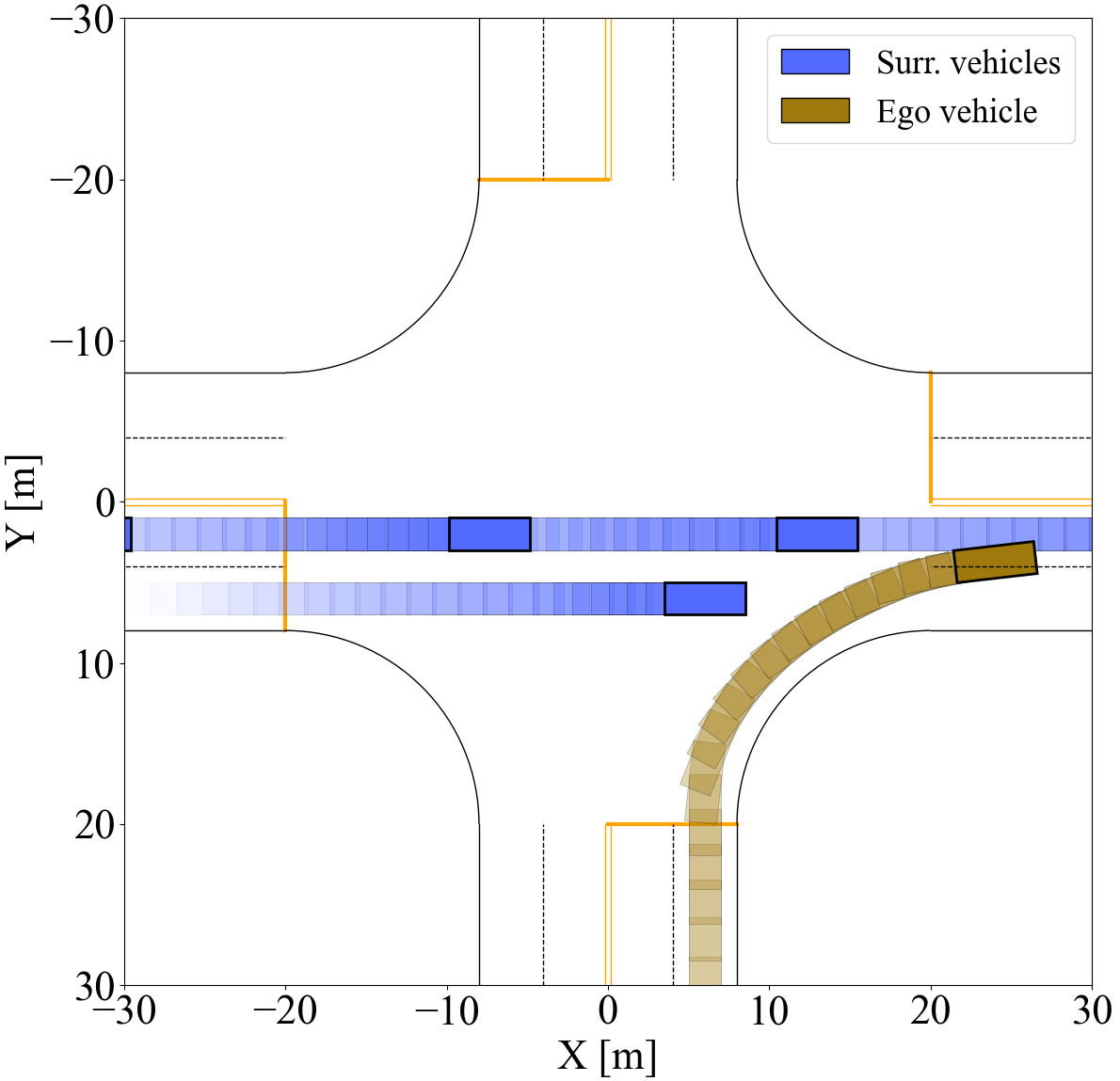}
        \label{right_sacrs}
    }\subfloat[]{%
        \includegraphics[width=0.19\textwidth]{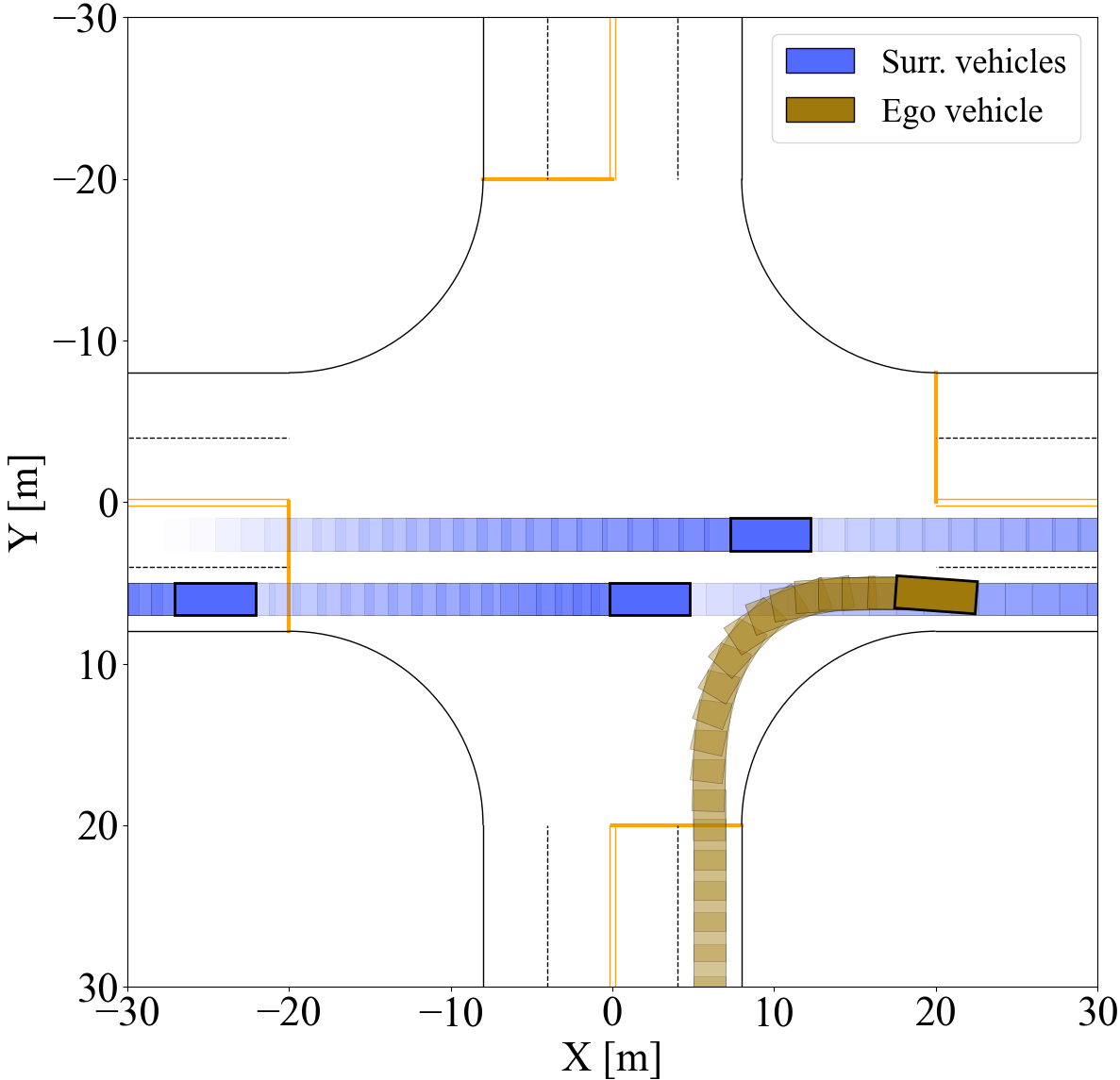}
        \label{right_ppors}
    }\subfloat[]{%
        \includegraphics[width=0.19\textwidth]{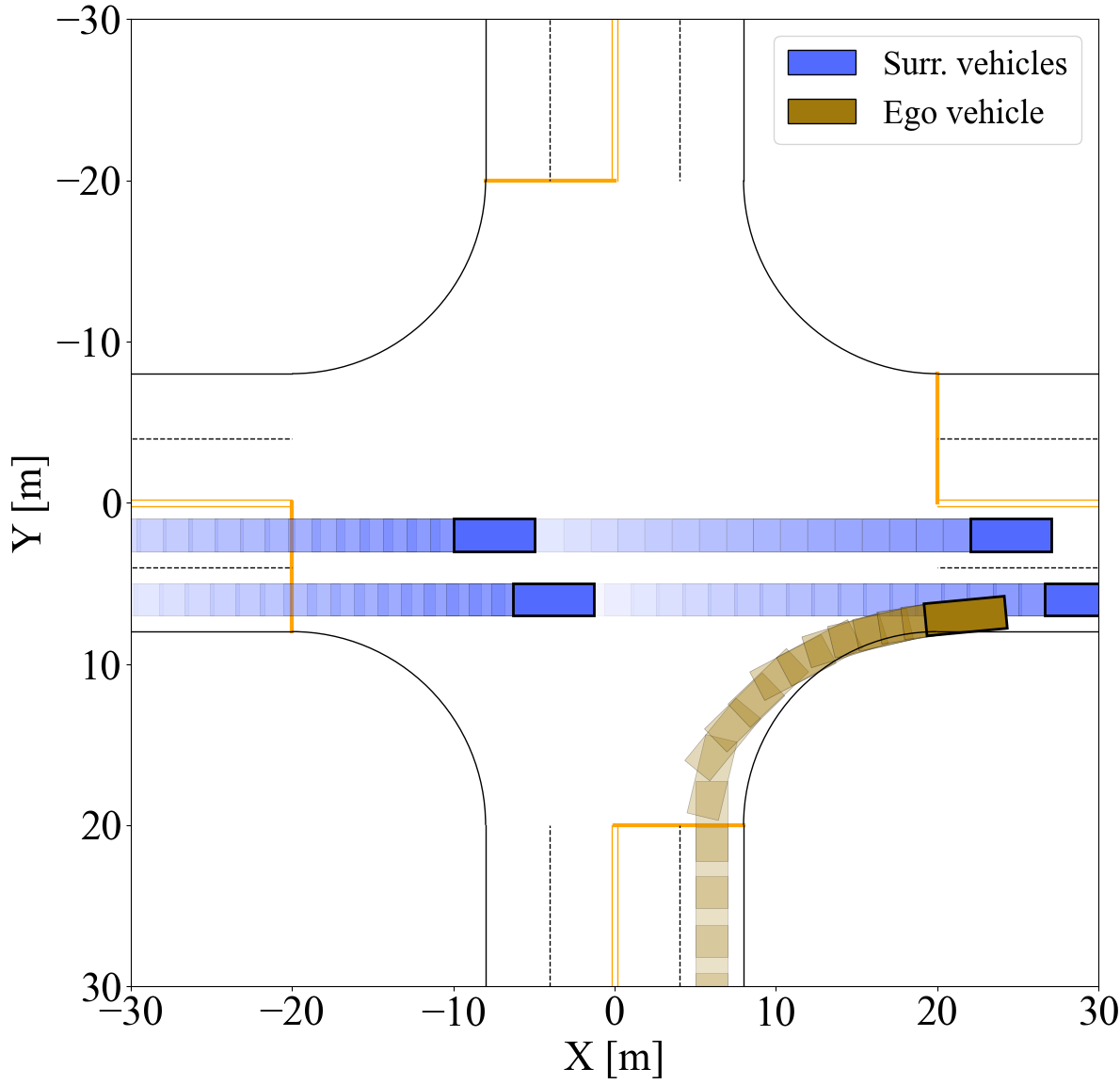}
        \label{right_cpo}
    }\\ \vspace{-5pt}
    \subfloat[]{%
        \includegraphics[width=0.98\textwidth]{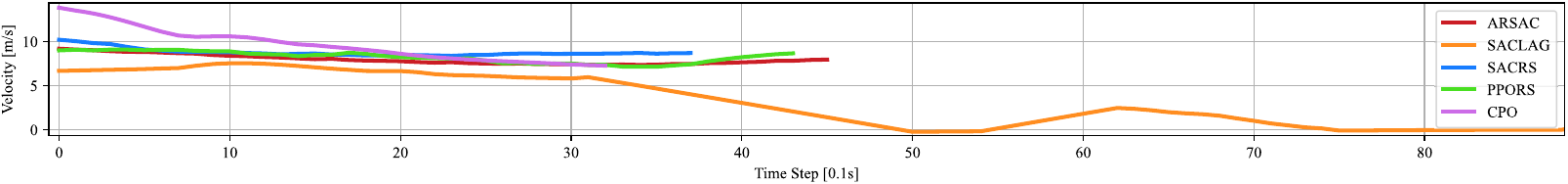}
        \label{right_velocity}
    }
    \caption{Vehicle trajectories visualization. Blue rectangles are SVs, brown rectangles are EVs. (a) ARSAC. (b) SAC-Lag. (c) SAC-RS. (d) PPO-RS. (e) CPO. (f) The velocity of the EV driving inside intersection.}
    \label{slice_compare_right}
\end{figure*}
\subsubsection{Right-turn Case}
In the RT case, all algorithms except SAC-Lag are able to produce safe, collision-free, and smooth trajectories, as depicted in Fig. \ref{slice_compare_right}. Fig. \subref*{right_velocity} illustrates that SAC-Lag attempts to merge into the traffic by reducing its speed. However, due to its inability to accurately gauge the optimal timing for merging, it adopts an overly cautious approach to avoid collisions, which ultimately reduces traffic efficiency. CPO enters the intersection at a speed of approximately 15 m/s. To avoid a collision, it rapidly decelerates to 10 m/s within about 0.7 seconds, then gradually reduces speed after 1.3 seconds to complete the merging maneuver. In contrast, ARSAC maintains a consistently stable speed throughout, highlighting its ability to accurately assess the optimal timing for merging while ensuring safety.

\section{Conclusion}
In this paper, we propose a risk-aware reinforcement learning algorithm to ensure that autonomous vehicles can safely and efficiently traverse intersection scenarios. Safe critics are designed to assess driving risks and work in conjunction with the reward critic to update the actor. Building on this, a Lagrangian relaxation method and cyclic gradient iteration are employed to project actions into a feasible safe region. Furthermore, a multi-hop, MLP-mixed attention mechanism is integrated into the actor-critic network, enabling the policy to adapt to dynamic traffic and overcome permutation sensitivity challenges, thereby allowing it to more effectively focus on surrounding potential risks while enhancing the identification of passing opportunities. Experimental results for left-turn, right-turn, and go straight driving tasks demonstrate that our algorithm effectively reduces collision rates and improves the efficiency of EV compared to baseline algorithms. Nonetheless, it is well known that autonomous vehicles share the road with various traffic participants, such as cyclists and pedestrians in the real-world environment. Therefore, our future work will extend to mixed traffic flows. In addition, we will use more accurate time series forecasting models and integrate predictive features to better assess risk. We also plan to use offline datasets  for training and testing, while extending the approach to a wider range of scenarios.


\bibliographystyle{elsarticle-harv}  

\bibliography{main}

\end{document}